\pdfoutput=1

\documentclass[11pt]{article}

\usepackage{amsmath,amsfonts,bm}

\def\eqref#1{equation~\ref{#1}}

\def\1{\bm{1}}

\def\rd{{\textnormal{d}}}

\def\rl{{\textnormal{l}}}

\def\rp{{\textnormal{p}}}

\def\rr{{\textnormal{r}}}

\DeclareMathAlphabet{\mathsfit}{\encodingdefault}{\sfdefault}{m}{sl}
\SetMathAlphabet{\mathsfit}{bold}{\encodingdefault}{\sfdefault}{bx}{n}

\def\sP{{\mathbb{P}}}

\def\sR{{\mathbb{R}}}

\usepackage[final]{acl}

\usepackage{times}
\usepackage{latexsym}

\usepackage[T1]{fontenc}

\usepackage[utf8]{inputenc}

\usepackage{microtype}

\usepackage{inconsolata}

\usepackage{graphicx}
\usepackage{booktabs}
\usepackage{multirow}
\usepackage{xspace}
\usepackage{enumitem}
\usepackage{bbding}
\usepackage{listings}
\usepackage[most]{tcolorbox}
\usepackage{tabularx}
\usepackage{url}
\usepackage{xcolor}
\newcommand{\model}{\texttt{SceneGenAgent}\xspace}
\newcommand{\dataset}{\texttt{SceneInstruct}\xspace}
\newcommand{\vpara}[1]{\noindent\textbf{#1}\xspace}
\definecolor{darkred}{RGB}{176,36,24}
\title{SceneGenAgent: Precise Industrial Scene Generation with Coding Agent}

\author{
Xiao Xia$^{1}$\thanks{\ \ Equal contribution.},
Dan Zhang$^{1}$\footnotemark[1],
Zibo Liao$^{2}$,
Zhenyu Hou$^{1}$,\\
\textbf{Tianrui Sun$^{2}$},
\textbf{Jing Li$^{2}$},
\textbf{Ling Fu$^{2}$},
\textbf{Yuxiao Dong$^{1}$} \\
$^{1}$Tsinghua University\quad$^{2}$Foundational Technologies, Siemens Ltd., China\\
}

\begin{document}
\maketitle

\begin{abstract}
The modeling of industrial scenes is essential for simulations in industrial manufacturing.
While large language models (LLMs) have shown significant progress in generating general 3D scenes from textual descriptions, generating industrial scenes with LLMs poses a unique challenge due to their demand for precise measurements and positioning, requiring complex planning over spatial arrangement.
To address this challenge, we introduce \model, an LLM-based agent for generating industrial scenes through C\# code. 
\model ensures precise layout planning through a structured and calculable format, layout verification, and iterative refinement to meet the quantitative requirements of industrial scenarios.
Experiment results demonstrate that LLMs powered by \model exceed their original performance, reaching up to 81.0\% success rate in real-world industrial scene generation tasks and effectively meeting most scene generation requirements.
To further enhance accessibility, we construct \dataset, a dataset designed for fine-tuning open-source LLMs to integrate into \model.
Experiments show that fine-tuning open-source LLMs on \dataset yields significant performance improvements, with Llama3.1-70B approaching the capabilities of GPT-4o.
Our code and dataset are available at \url{https://github.com/THUDM/SceneGenAgent}.
\end{abstract}

\section{Introduction}
\begin{figure*}[t!]
  \centering
  \includegraphics[width=0.98\textwidth]{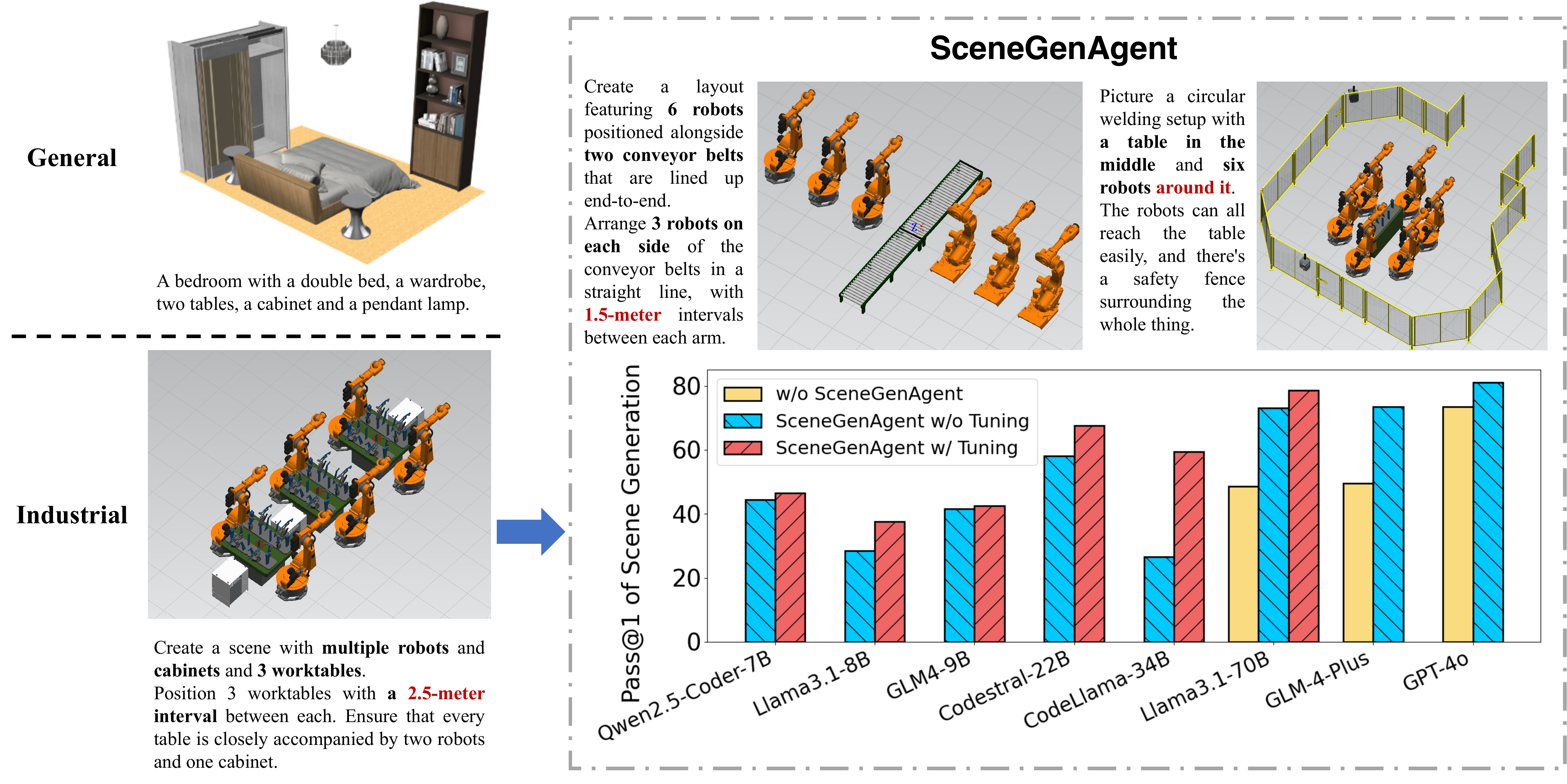}
  \caption{\textbf{Left: General and industrial scene generation examples.} Industrial scene generation demands quantitative precision inputs compared to general scenes. \textbf{Right: Examples of \model and leaderboard of LLMs incorporated in \model for industrial scene generation.}
  Llama3.1-70B and API-based LLMs achieve high $\text{pass}@1$.
  Fine-tuned LLMs show consistent improvement for different parameter sizes. General scene example is selected from LayoutGPT~\cite{feng2024layoutgpt} and industrial scene examples are generated by \model with the fine-tuned Llama3.1-70B as its backbone.}
  \label{fig:intro-examples}
\end{figure*}

The advancement of large language models (LLMs)~\citep{brown2020language, kaplan2020scaling, touvron2023llama, zengglm, glm2024chatglm} enables the automatic design of scene layouts, providing a promising direction for the automatic generation of industrial scenes.
Automatically generated digital modeling of industrial scenes can be used for design visualization, simulation, analysis, and optimization in manufacturing.
Recent scene generation research~\citep{sun20233d, yang2024holodeck, hu2024scenecraft} has made remarkable achievements in translating textual descriptions into 3D scenes, grasping a delicate understanding of spatial layout.

However, compared to general scene generation, the datasets and evaluations of which~\citep{chang2017sceneseer, fu20213d, sun20233d, de2024llmr, yang2024holodeck} often prioritize qualitative plausibility, \textit{industrial scene generation} presents the challenge of requiring quantitative measurements and precise positioning to maintain the functional integrity of machinery and safety compliance.
Industrial scenes are described with specified design parameters and constraints such as coordinates, distances, and orientations to which the components must align with strict adherence (cf. Figure~\ref{fig:intro-examples}).
However, current LLMs struggle to meet the precision requirements of industrial scene generation without sufficient quantitative analysis of complex layout~\citep{yang2024llm}, and we also demonstrate this drawback in Table~\ref{tab:exp-abl-layout}.
Meanwhile, LLMs used as agents~\citep{shinn2023reflexion, yao2024tree, wang2023describe, xu2024androidlab, wang2024battleagentbench, zhang2025datascibench} bring possibilities to handle complex scenes with planning.

In this paper, we introduce \model, an LLM-based agent that enables precise generation of industrial scenes.
\model translates scene descriptions into C\# code, leveraging APIs provided by Tecnomatix\footnote{https://plm.sw.siemens.com/en-US/tecnomatix/}, a dedicated digital manufacturing software developed by Siemens, and executes the code to render the corresponding scene.
We prioritize the accurate placement within the scene in \model by designing a layout analysis pipeline.
Specifically, \model \textbf{plans the overall layout} by representing spatial specifications including positions and relations in a structured, calculable format and computing and reasoning with numerical constraints.
After producing a layout plan, \model \textbf{verifies the layout} for compliance with the description and employs identified flaws as feedback to refine the layout.
Finally, \model \textbf{generates code} based on the layout plan and Tecnomatix API coding guidance.
Experimental results show that LLMs powered by \model exceed their original performance, with GPT-4o~\citep{achiam2023gpt} reaching 81.0\% success rate in real-world industrial scene generation tasks and effectively meeting most scene generation requirements.

We introduce an approach for fine-tuning open-source LLMs to integrate into \model, allowing individuals to deploy it independently. Given the absence of publicly available datasets for industrial scenes, we construct \dataset, a dataset capturing the complexity of industrial scene descriptions and covering key steps in \model pipeline.
Experiments show that fine-tuned LLMs of different scales exhibit significant improvements over their counterparts without tuning, with Llama3.1-70B~\cite{dubey2024llama} nearing the performance of the leading GPT-4o, showcasing the efficacy of our approach.

\begin{figure*}[t!]
  \centering
  \includegraphics[width=\textwidth]{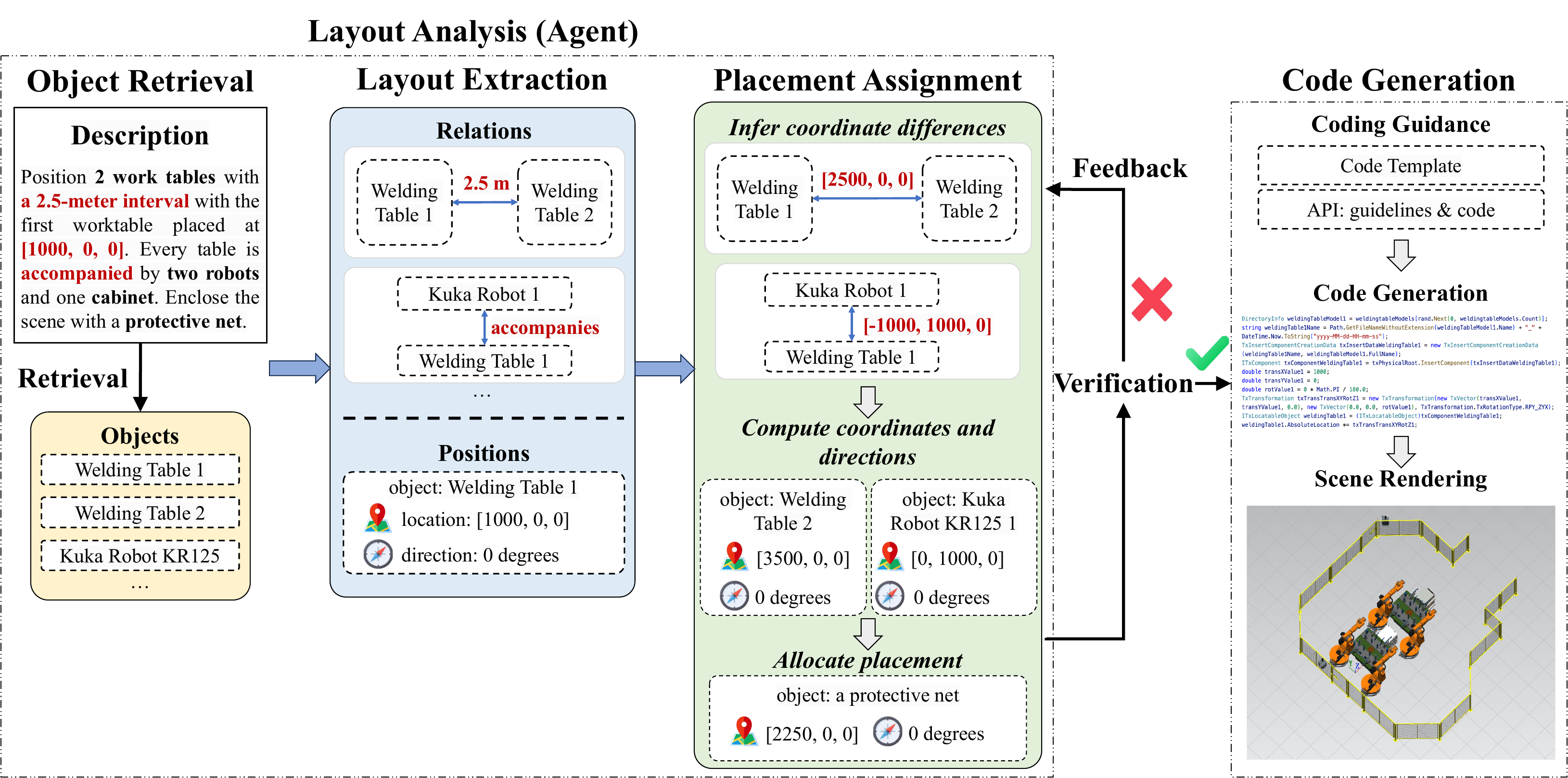}
  \caption{\textbf{Overview of \model framework.} Based on the description, the agent first retrieves objects from the library and then extracts positions and relations from the description. The relations are converted into coordinate differences, which are used together with the positions to calculate and infer the placement of all objects. After passing verification, the placement is used to generate code.
  }
  \label{fig:method-overview}
\end{figure*}

The contributions are summarized as follows:
\begin{itemize}[leftmargin=*,itemsep=0pt,parsep=0.5em,topsep=0.3em,partopsep=0.3em]
    \item We propose \model, an LLM-based coding agent capable of generating industrial scenes from descriptions, analyzing and improving layout for precise generation, shown in Figure~\ref{fig:method-overview}.
    
    \item We construct \dataset, a dataset that encompasses the core functions of \model and is used to fine-tune open LLMs to enable individual agent deployment, detailed in Table~\ref{tab:method-dataset}.
    
    \item Our experiments demonstrate that API-based and open-source fine-tuned LLMs empowered by \model achieve high success rates in real-world industrial scene generation tasks, as summarized in Table~\ref{tab:exp-main} and Table~\ref{tab:exp-abl-layout}.
\end{itemize}

\section{Preliminary}
\paragraph{Modeling Industrial Scenes.}
A scene can be represented as a collection of 3D models, denoted as objects $\mathcal{O}$, drawn from an object library.
Each object $o_i \in \mathcal{O}$ in the scene is characterized by its coordinates $c_i$ and direction $d_i$.
The coordinates specify the object's location on a two-dimensional plane, where the positive x-axis represents the front and the positive y-axis represents the left.
The direction defines the object's angle rotated counterclockwise from the positive direction of the x-axis.
The placement of an object, $p_i = (o_i, c_i, d_i)$, encapsulates its spatial properties.
The Euclidean distance between two objects must be at least 1 meter to avoid collision.
The set of all objects' placements $\mathcal{P}=\{p_i=(o_i, c_i, d_i)|o_i\in\mathcal{O}\}$ provides a modeling of the scene.
Given an industrial scene's description, we aim to identify the objects and their placements to reconstruct the scene.

\paragraph{Code Representation of Industrial Scenes.}
We use C\# code to leverage Tecnomatix APIs and model industrial scenes.
The code aims to place objects from $\mathcal{O}$ according to the predefined placement $\mathcal{P}$.
We restrict the code's functionality to loading object models and placing objects to streamline API integration.
The first selects and loads the model files of objects according to $\mathcal{O}$, and the second sets coordinates and directions for objects with specifications outlined in $\mathcal{P}$.
The code is run in the Process Simulate software\footnote{https://plm.sw.siemens.com/en-US/tecnomatix/products/process-simulate-software/} to render the scene.

\section{Method}
This section consists of two parts: \textbf{\model} and \textbf{\dataset}.
Regarding the first part, we design an agent, \model, as shown in Figure~\ref{fig:method-overview}, which leverages LLMs to translate the description into scene-modeling code.
In the second part, we synthesize a new dataset \dataset to fine-tune open-source LLMs to achieve performance comparable to API-based models.

\subsection{\model}
We introduce the workflow of an industrial scene-generating agent, which consists of two phases: \textbf{Layout Analysis (Agent)} aiming to extract and analyze the scene's layout and \textbf{Code Generation} for generating code using Tecnomatix APIs.

\subsubsection{Layout Analysis (Agent)}
\model performs layout analysis to recover the objects $\mathcal{O}$ and placements $\mathcal{P}$.
Layout analysis consists of three stages: object retrieval, layout extraction, and placement assignment.

\paragraph{Object retrieval.}
\model identifies the objects in the scene and retrieves the corresponding items from the object library.
The agent executes all the following operations in a single LLM call using the description as input.

First, the agent generates a list of objects that should appear in the scene based on the description.
This list includes ambiguously mentioned objects like ``a random object''.
The agent then maps the objects to their counterparts within the library by incorporating the objects from the library and objects found in the description into the LLM's input context and tasking the LLM to verify the presence of each object in the description within the library.
If an object is not found in the library, the agent attempts to substitute it with a similar object from the library that conforms to the description.
In cases where no suitable substitute is available, the agent removes the object.
Consequently, the agent retrieves the object collection $\mathcal{O}=\{o_i\}$ from the library.
Finally, the agent rewrites the description to reflect the retrieved objects' layout, replacing the original objects' names with those of the retrieved ones and retaining solely layout information.
The output of this rewriting is the layout's concise depiction, denoted as the new description $s$.

\paragraph{Layout extraction.}
\model extracts all layout information from $s$.
We categorize the existing layout information in $s$ into two types: 
\begin{itemize}[leftmargin=*,itemsep=0pt,parsep=0.5em,topsep=0.3em,partopsep=0.3em]
    \item \textbf{Position}, which refers to the positional state of an object, noted as $\rp=(o_i, \rl_i, \rd_i)$, where $o_i$ denotes the object, $\rl_i$ the location, and $\rd_i$ the direction.
    Note that some objects may have no $\rp$ or their $\rl_i$ or $\rd_i$ may be missing because the description may not provide this information.
    \item \textbf{Relation}, which concerns the interactions between two objects, including aspects such as distance and relative direction.
    This is expressed as $\rr=(o_i, o_j, r_{ij})$, where $o_i$ and $o_j$ denotes the two objects and $r_{ij}$ their relationship.
    $r_{ij}$ is a text describing relationships such as adjacency, parallel, surrounding, etc.
    Relations involving multiple objects are represented as a series of relations between object pairs, which, accompanied by the description $s$, can be comprehended by the agent.
\end{itemize}
The existing layout information in $s$ can be formalized into two sets $\sP=\{\rp\}$ and $\sR=\{\rr\}$, representing the set of positions and relations respectively.
The agent extracts all layout information with an LLM call, incorporating $s$ and $\mathcal{O}$ into the input and listing positions and relations in the output.

\paragraph{Placement assignment.}
\model assigns the final placement of every object $\mathcal{P}$ based on $\sP$ and $\sR$ and improves $\mathcal{P}$ through verification.
Beyond adhering to the description, the assignment must also satisfy supplementary constraints, i.e., the specifications concerning object distance and orientation, to ensure the coherence of the scene.
The agent undertakes the following steps:
\begin{enumerate}[leftmargin=*,itemsep=0pt,parsep=0.5em,topsep=0.3em,partopsep=0.3em]
    \item \textit{Infer placement from $\sP$}: For each position information $\rp=(o_i, \rl_i, \rd_i)$, if $c_i$ or $d_i$ of $o_i$ is included in or can be deduced from $\rl_i$ or $\rd_i$, assign $c_i$ or $d_i$ to $o_i$.
    \item \textit{Infer coordinate differences from $\sR$}: Relations $\rr=(o_i, o_j, r_{ij})\in\sR$ are viewed as differences in coordinates between two objects. For instance, the distance between two objects is expressed as the magnitude of their coordinate difference vector.
    Rewrite all relations into coordinate differences, forming the set $\Delta=\{(o_i, o_j, \delta_{ij})|\delta_{ij}\text{ is rewritten from }r_{ij},(o_i, o_j,\allowbreak r_{ij})\in\sR\}$, where $\delta_{ij}$ represents the vector difference between the coordinates of $o_i$ and $o_j$.
    \item \textit{Compute coordinates and directions}: For each coordinate difference $(o_i, o_j, \delta_{ij})\in\Delta$, if the coordinates of one object are known, compute the other's coordinates based on $\delta_{ij}$. For each relation $\rr=(o_i, o_j, r_{ij})\in\sR$, if $r_{ij}$ states that $o_i$ is oriented towards $o_j$ and the coordinates of both objects are known, calculate the direction of $o_i$, and vice versa. 
    \item \textit{Allocate placement to the rest of the objects}: Allocate placement to objects whose coordinates or directions have not yet been determined without collision, adhering to the description.
\end{enumerate}
All objects are placed after these steps, forming the initial placement $\mathcal{P}$.

\vpara{Verification.}
The initial placement may contain flaws since the LLM may make common-sense errors or miscalculate values.
Recent studies~\citep{shinn2023reflexion, madaan2024self} have shown the ability of language models to critique their own decisions.
Hence, \model iteratively checks $\mathcal{P}$ for the following potential errors:
(1) whether the placement violates the constraints described in $s$ and
(2) whether any objects collide, measured by the distances between adjacent objects.
If errors are identified, the agent uses $\sP$, $\sR$, $\mathcal{P}$, and the detected errors as input and attempts to fix the errors to generate a new $\mathcal{P}$ until no errors are found.

\subsubsection{Code Generation}
This phase involves writing C\# code with Tecnomatix APIs to visualize the layout and model the scene.
We design a code template containing basic code, such as initialization, and leave the parts responsible for model loading and object placement incomplete for the agent to fill.
Following prior works~\citep{zhoudocprompting} that integrate code documentation into the model's input, we provide \textbf{coding guidance} alongside the template, aiding the agent in using APIs.
The coding guidance includes usage guidelines and example code snippets corresponding to APIs.
The agent employs an LLM to \textbf{generate code}, integrating $\mathcal{P}$, code template, and coding guidance into the input.

Potential errors can be identified within the generated code, including omitting code or making up methods or variables unsupported by Tecnomatix.
The agent conducts a rule-based evaluation by scanning keywords associated with the APIs in the code to assess whether the LLM has correctly implemented the API calls.
The detected invalid code is used as feedback to prompt the LLM to revise the code accordingly.
Once the code passes the check, we run it to \textbf{render the industrial scene}.

\subsection{\dataset}
This section details our efforts in training open-source LLMs to integrate into \model.
Inspired by previous research~\citep{zeng2023agenttuning, chen2023fireact} fine-tuning LLMs on agent trajectories to enhance their agent abilities, we gather the agent's trajectory data to build \dataset, constructing a collection of industrial scene descriptions and feeding them into the agent with API-based models as its backbone.
Then, we use \dataset to fine-tune open-source LLMs.

\subsubsection{\dataset Construction}
\label{sec:dataset-construction}
\paragraph{Instruction Construction.}
We construct a description dataset that serves as the agent's instruction to obtain its behavior trajectory.
We manually write 20 scene descriptions as the initial description pool, encompassing fundamental variations of industrial scenes.
Then, we adopt the Evol-Instruct method~\citep{xu2024wizardlm} to rewrite descriptions iteratively with an LLM, generating diverse variants.
We design 6 rewriting methods based on possible variations of descriptions.
In each iteration, a random description from the current pool and a rewriting method are sampled.
The description is rewritten by Llama3.1-70B~\citep{dubey2024llama} into a new description.
Appendix~\ref{app:instruct-construct-detail} details the rewriting strategies and sampling.

Each new description undergoes validation and deduplication.
Validation detects errors including common-sense mistakes like collisions and contradictions between different parts of the description.
An object's placement might be specified multiple times, but its placements derived in these different ways may not match.
The LLM first attempts to compute placements, distances, and relative positions as much as possible based on the description, then examines common-sense errors and conflicts in calculated results.
Deduplication evaluates the similarity between new and existing descriptions with the MinHash~\citep{broder2000identifying} algorithm.
The LLM retries rewriting if validation or deduplication fails, and then we add it to the description pool for sampling in later iterations.
The final dataset includes 3002 descriptions, from which we use 95\% (2852) for training and the rest for validation.

\paragraph{Trajectory Collection.}
After constructing the instructions, we collect \model's trajectories with GPT-4~\citep{achiam2023gpt} as its backbone, gathering the inputs and outputs of GPT-4 within the workflow.
We refer to Llama3.1-70B, given its capabilities for instruction following, multi-turn dialogue, and writing C\# code, to select which parts of the agent need data collection.
We observe that while it can handle most operations, the performance in assigning placement can be improved.
Therefore, we gather data from this stage, where the LLM should complete three tasks: assign, verify, and reassign placement.
For each description, Llama3.1-70B first completes the agent's tasks up to extracting layout, then GPT-4 assigns placement and we collect its inputs and outputs.

Data containing inaccuracies is crucial to train the model to be sensitive to errors during placement verification.
Nevertheless, the positions assigned by GPT-4 tend to be classified as correct.
Thus, we add extra data by changing the LLM that assigns placement to the weaker Llama3.1-8B to obtain more data with errors.
We add this data to the existing dataset, forming \dataset.
Table~\ref{tab:method-dataset} shows the statistics of \dataset for training.

\begin{table}[t!]
  \centering
  \caption{\textbf{Overview of \dataset for training.} \textbf{Assign} represents assigning, \textbf{Verify} represents verifying whereas (Pos/Neg) represents data with/without errors, and \textbf{Reassign} represents reassigning. ``Base'' represents data constructed with GPT-4 and ``Extra'' represents data constructed with Llama3.1-8B assigning placement.}
  \resizebox{!}{12mm}{
  \begin{tabular}{lccc}
    \toprule
    \textbf{Task} & \textbf{Assign} & \textbf{Verify (Pos/Neg)} & \textbf{Reassign} \\
    \midrule
    \#Base & 2852 & 5413 (2561/2852) & 682 \\
    \#Extra & 2852 & 6984 (4132/2852) & 1905 \\
    \#Total & 5704 & 12397 (6693/5704) & 2587 \\
    \bottomrule
  \end{tabular}
  }
  \label{tab:method-dataset}
\end{table}

\begin{table*}[t!]
  \centering
  \caption{\textbf{Main results.} The best performances for open or closed LLMs are in \textbf{bold}. All LLMs are evaluated by incorporating them into the \model. API-based LLMs complete all agent operations, and open LLMs are utilized during placement assignment while Llama3.1-70B handles the remaining operations. For $\text{pass}@1$ categories, Geo. represents geometric arrangement, Pos. represents positional details, Quant. represents object quantity, Comp. represents composite description, Fuzz. represents fuzzy description, and Overall represents the overall metric.}
  \resizebox{1.86\columnwidth}{!}{
  \begin{tabular}{c|c|c|cccccc}
    \toprule
    \multirow{1}{*}{\textbf{Model}} &\multirow{2}{*}{\textbf{Type}} & \multirow{2}{*}{\textbf{Fine-tuned}} & \multicolumn{6}{c}{\textbf{$\text{pass}@1$}} \\
    \cmidrule(lr){4-9}
     (w/ \model) & & & Geo. & Pos. & Quant. & Comp. & Fuzz. & Overall \\
    \midrule
    GLM-4-Plus & closed & -  & 72.0 & 68.9 & \textbf{76.7} & 68.0 & \textbf{90.0} & 73.5 \\
    GPT-4o & closed & -  & \textbf{81.3} & \textbf{82.2} & 73.3 & \textbf{84.0} & \textbf{90.0} & \textbf{81.0} \\
    \midrule
    \multirow{2}{*}{Qwen2.5-Coder-7B} & \multirow{2}{*}{open} & \XSolidBrush  & 40.0 & 42.2 & 46.7 & 32.0 & 80.0 & 44.5 \\
     & & \Checkmark & 53.3 & 33.3 & 53.3 & 40.0 & 60.0 & 46.5 \\
     \cmidrule(lr){1-9}
    \multirow{2}{*}{Llama3.1-8B} & \multirow{2}{*}{open} & \XSolidBrush  & 16.0 & 15.6 & 40.0 & 24.0 & 90.0 & 28.5 \\
     & & \Checkmark  & 22.7 & 33.3 & 46.7 & 44.0 & 85.0 & 37.5 \\
     \cmidrule(lr){1-9}
    \multirow{2}{*}{GLM4-9B} & \multirow{2}{*}{open} & \XSolidBrush  & 33.3 & 35.6 & 46.7 & 48.0 & 75.0 & 41.5 \\
     & & \Checkmark  & 36.0 & 26.7 & 50.0 & 44.0 & 85.0 & 42.5 \\
     \cmidrule(lr){1-9}
    \multirow{2}{*}{Codestral-22B} & \multirow{2}{*}{open} & \XSolidBrush  & 54.7 & 46.7 & 70.0 & 48.0 & 90.0 & 58.0 \\
     & & \Checkmark  & 62.7 & 62.2 & \textbf{80.0} & 52.0 & 90.0 & 67.5 \\
     \cmidrule(lr){1-9}
     \multirow{2}{*}{CodeLlama-34B} & \multirow{2}{*}{open} & \XSolidBrush  & 26.7 & 17.8 & 26.7 & 28.0 & 50.0 & 26.5 \\
     & & \Checkmark  & 50.7 & 55.6 & 60.0 & \textbf{64.0} & 85.0 & 59.5  \\
     \cmidrule(lr){1-9}
    \multirow{2}{*}{Llama3.1-70B} & \multirow{2}{*}{open} & \XSolidBrush  & 64.0 & \textbf{80.0} & 76.7 & \textbf{64.0} & \textbf{95.0} & 73.0 \\
     & & \Checkmark  & \textbf{78.7} & 77.8 & \textbf{80.0} & 60.0 & 90.0 & \textbf{78.5} \\
    \bottomrule
  \end{tabular}
  }
  \label{tab:exp-main}
\end{table*}

\subsubsection{\dataset Tuning}
\label{sec:data-tuning}
We choose the chat versions of Qwen2.5-Coder-7B~\citep{hui2024qwen2}, Llama3.1-8B~\citep{dubey2024llama}, GLM4-9B~\citep{glm2024chatglm}, Codestral-22B~\citep{jiang2023mistral}, CodeLlama-34B~\citep{roziere2023code}, and Llama3.1-70B~\citep{dubey2024llama} as base models for fine-tuning.
For a base model, we utilize LoRA~\citep{hu2022lora}, fine-tune multiple LoRA modules corresponding to tasks, and dynamically switch modules in deployment.
We organize data for assignment and verification as question-answer pairs and data for reassigning as conversations.
We calculate the loss of outputs in the final rounds during fine-tuning.
Detailed training hyper-parameters are in Appendix~\ref{app:train-hyperparams}.
\section{Experiments}
\subsection{Experiment Setup}

\paragraph{Benchmark.} We develop a comprehensive industrial scene layout generation benchmark containing 40 descriptions of industrial scenes.
These descriptions are written by 15 engineers based on their experience and split into 5 categories: geometric arrangement, focusing on geometric patterns; positional details, dealing with numerical measurements; object quantity, about the number of objects present; composite description, involving detailed, long-form textual descriptions; fuzzy description, an overview without precise specifications.
The proportion of each category reflects its frequency of occurrence.
Details are provided in Appendix~\ref{app:benchmark}.

\begin{table*}[t!]
  \centering
  \caption{\textbf{Ablation study on the effect of layout analysis.} ``w/o agent'' indicates obtaining objects and layout in one LLM call, ``w/o verification'' indicates removing verification during placement assignment, and ``w/ \model'' indicates the designed workflow. For $\text{pass}@1$ categories, Geo.\& Pos. represents combining geometric arrangement and positional details, and the remaining categories are the same as Table~\ref{tab:exp-main}.}
  \begin{tabular}{c|l|ccccc}
    \toprule
    \multirow{2}{*}{\textbf{Model}} & \multirow{2}{*}{\textbf{Layout Analysis}} & \multicolumn{5}{c}{\textbf{$\text{pass}@1$}} \\
    \cmidrule(lr){3-7}
     & & Geo. \& Pos. & Quant. & Comp. & Fuzz. & Overall \\
    \midrule
    \multirow{3}{*}{GPT-4o} & w/ \model & 81.7 & 73.3 & 84.0 & 90.0 & 81.0 \\
     & w/o agent & 72.5 & 83.3 & 64.0 & 90.0 & 73.5 (\textbf{\textcolor{darkred}{-9.3\%}}) \\
     & w/o verification & 73.3 & 80.0 & 76.0 & 95.0 & 77.0 (\textbf{\textcolor{darkred}{-4.9\%}}) \\
    \midrule
    \multirow{3}{*}{GLM-4-Plus} & w/ \model & 70.8 & 76.7 & 68.0 & 90.0 & 73.5 \\
     & w/o agent & 46.7 & 60.0 & 52.0 & 60.0 & 49.5 (\textbf{\textcolor{darkred}{-32.7\%}}) \\
     & w/o verification & 58.3 & 56.7 & 44.0 & 90.0 & 60.5 (\textbf{\textcolor{darkred}{-17.7\%}}) \\
    \midrule
    \multirow{3}{*}{Llama3.1-70B} & w/ \model & 70.0 & 76.7 & 64.0 & 95.0 & 73.0 \\
     & w/o agent & 44.2 & 43.3 & 40.0 & 80.0 & 48.5 (\textbf{\textcolor{darkred}{-33.6\%}}) \\
     & w/o verification & 70.0 & 73.3 & 60.0 & 80.0 & 70.5 (\textbf{\textcolor{darkred}{-3.4\%}}) \\
    \midrule
    \multirow{2}{*}{\begin{tabular}[x]{@{}l@{}}Llama3.1-70B\\(Fine-tuned)\end{tabular}} & w/ \model & 78.3 & 80.0 & 60.0 & 90.0 & 78.5 \\
     & w/o verification & 69.2 & 76.7 & 68.0 & 85.0 & 72.5 (\textbf{\textcolor{darkred}{-7.6\%}}) \\
    \bottomrule
  \end{tabular}
  \label{tab:exp-abl-layout}
\end{table*}

\paragraph{Metric and evaluation.} We adopt the $\text{pass}@k$ metric~\citep{chen2021evaluating}, a commonly used metric for coding performance that measures the ratio of programming problems where at least one out of $k$ generated code samples qualifies.
In the absence of automated testing tool suites, we rely on manual verification, carefully comparing each scene against its description.
Our evaluation team consists of 3 members with expertise in industrial scenes and proficiency in the Process Simulate software.
Due to the labor-intensive nature of the manual inspection, we sample 5 times for each description with temperature $t=1$ and evaluate $\text{pass}@1$ with the estimator $\text{pass}@k=\mathbb{E}[1-{\tbinom{n-c}{k}}/{\tbinom{n}{k}}]$, where $n$ is the number of samples for each problem (5 in this work), $c$ is the number of correct samples, and $k$ is set to 1.
Each evaluator reviews at least one of the five samples of each description to minimize the influence of subjectivity, mitigating potential biases introduced by different evaluators.

\paragraph{Models.} We evaluate the performance of multiple LLMs in \model.
For API-based LLMs, we select GPT-4o~\citep{achiam2023gpt} and GLM-4-Plus~\citep{glm2024chatglm}, both known for their strong coding and reasoning capabilities.
As for open-source LLMs, we include those we fine-tuned with \dataset in Section~\ref{sec:data-tuning}.
When incorporated into \model, API-based LLMs handle all agent operations, and open-source LLMs are applied during placement assignment with Llama3.1-70B managing the remaining operations.

\subsection{Main Results}
Table~\ref{tab:exp-main} presents the performance of LLMs integrated into the agent.
Larger models achieve high accuracy in the scene generation task, with GPT-4o exhibiting the highest $\text{pass}@1$ among all models, effectively solving most tasks.
The open models' $\text{pass}@1$ generally remain lower than closed models.
Smaller models solve only a limited subset of tasks, whereas Llama3.1-70B achieves performance comparable to closed models.
The results also suggest that fine-tuning with \dataset significantly benefits open models.
After fine-tuning, all open models improve overall $\text{pass}@1$, with most models also exhibiting higher $\text{pass}@1$ in most categories.
Llama3.1-70B approaches the top-performing GPT-4o, increasing its $\text{pass}@1$ from 73.0\% to 78.5\%.

Overall, the results highlight the efficacy of \model supported by closed and open LLMs for industrial scene generation.
Furthermore, the findings emphasize the effectiveness of our proposed data construction and fine-tuning methodology, with fine-tuning identified as a critical factor in enhancing the performance of open LLMs.

\subsection{Ablation Study}
We conduct detailed ablation studies to demonstrate the effectiveness of our approach in addressing key challenges in industrial scene generation, including the precise placement of objects and the limited availability of training data.

\subsubsection{Necessity of agent integration}
We implement a sophisticated agent workflow to detect objects within the scene, assign spatial placements, and verify and adjust them.
To confirm the necessity of this agent design, we test two settings:
(1) obtaining objects and layout in a single LLM call, that is, not adopting the agent methodology, 
and (2) omitting verification after assigning placement, simplifying the agent.
In the first case, the LLM should directly identify the objects and layout from the description.
We also add all requirements to the input, including object library and specifications concerning object distance and orientation, to guarantee that the LLM can access the same knowledge as the agent.
The prompt for this call is in Appendix~\ref{app:prompt-layout-analysis-single-call}.
In the second case, we skip verifying placement and directly proceed to code generation.
We evaluate the performance under both settings and compare it with the original \model.

We evaluate GPT-4o, GLM-4-Plus, and both base and fine-tuned Llama3.1-70B.
Since the fine-tuned Llama3.1-70B is trained with prompts different from those in the first setting, we only evaluate it in the second setting.
The results, provided in Table~\ref{tab:exp-abl-layout}, show performance degradation in all models without the agent or verification.
The performance drop in geometric arrangement and positional details marks the incidence of spatial errors rising, proving that the simplified settings lack the complexity required for precise scene generation.

\begin{figure}[t!]
  \centering
  \includegraphics[width=0.48\textwidth]{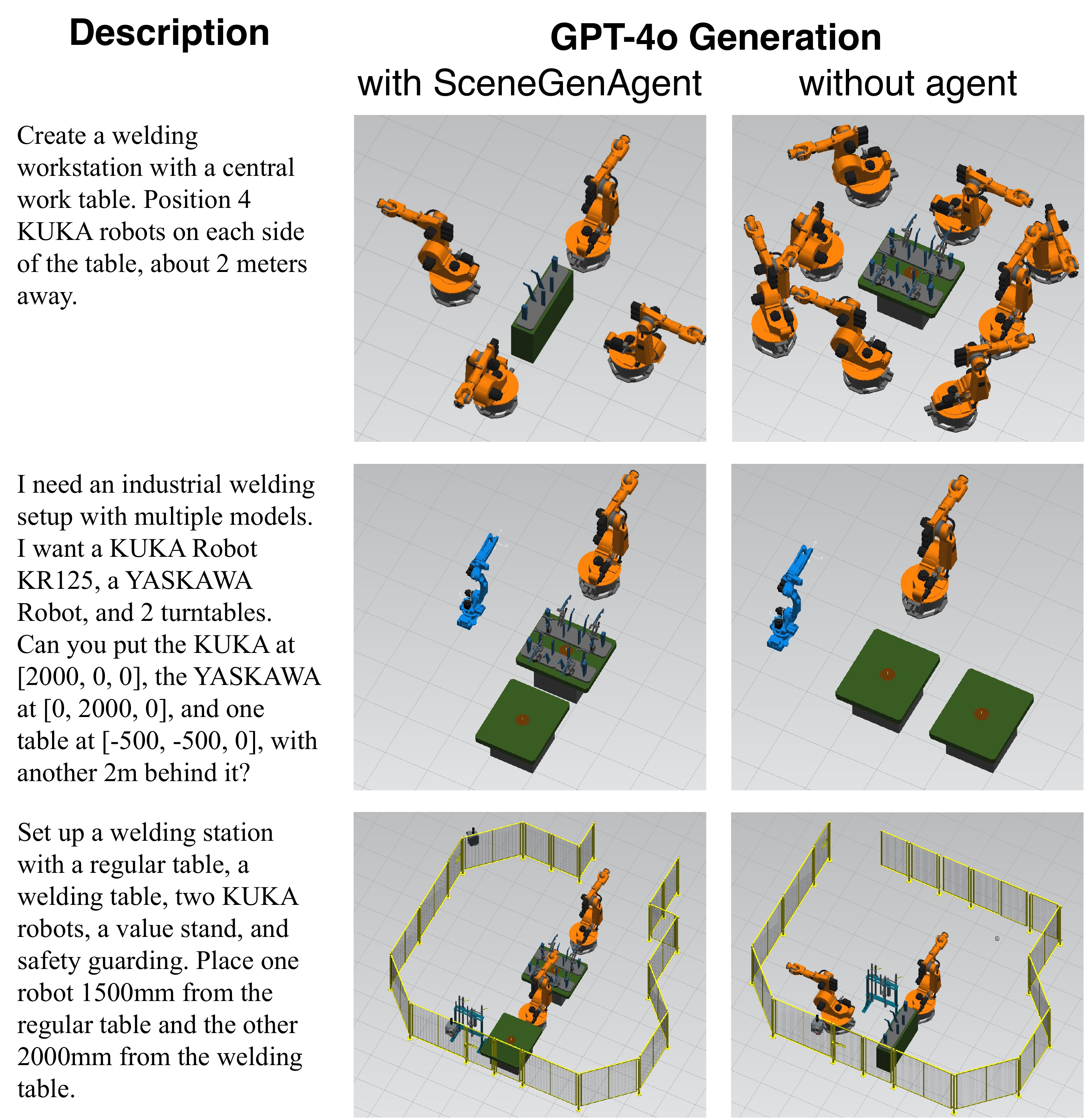}
  \caption{\textbf{Examples of scenes generated by GPT-4o with and without \model.}}
  \label{fig:case-study}
\end{figure}

To demonstrate the improvements achieved by powering GPT-4o with \model, we present several examples of scenes generated by GPT-4o, with and without the integration of \model, in Figure~\ref{fig:case-study}.
In the examples, all scenes generated by GPT-4o with \model accurately reflect the descriptions, whereas those generated without \model exhibit various inaccuracies.
Specifically, in the scenes generated without \model, the first example includes eight tables instead of the described four; the second misplaces the second table to the right of the first rather than behind it; and the third erroneously contains only one table instead of two and assumes the two robots are positioned relative to the same table rather than distinct ones.
These examples highlight that even advanced models like GPT-4o can be susceptible to errors concerning the number of objects, their spatial orientations, and inter-object relationships.
In contrast, \model, providing a nuanced understanding of placement, contributes notably to complex scene generation ability, even improving the performance of GPT-4o, which already has strong capabilities.

\subsubsection{Necessity of guidance in code generation}
\label{sec:guidance-code-gen}
During the code generation phase, we provide the LLM with coding guidance related to Tecnomatix APIs.
We assess whether the models require targeted guidance to utilize the APIs.
We experiment by removing the API-specific instructions for object placement and instead instructing the LLMs to write Tecnomatix-based code to set the coordinates and directions of objects, drawing on their inherent knowledge regarding the Tecnomatix framework.
This allows us to evaluate the models' performance without explicit coding guidance.
We present this prompt for code generation in Appendix~\ref{app:prompt-code-gen-wo-placement-guidance}.

We conduct these tests on GPT-4o and Llama3.1-70B without fine-tuning.
\textbf{All generated code fails to compile without API guidance, yielding 0 in $\text{pass}@1$ for both models.}
Rather than invoking the correct methods, the models often resort to fabricated methods not supported by Tecnomatix.
This finding reveals that LLMs struggle to produce functional code without framework-specific instruction and validates our approach, emphasizing the necessity of API guidance for generating code.

\subsubsection{Influence of data scaling}
Due to the limited availability of high-quality, domain-specific data in industrial applications, we investigate the impact of the quantity of training data on performance to explore how much data is needed for scene generation capability~\citep{zhang2024sciglm,zhang2024rest}.
We conduct experiments using varying proportions of the original training data and assess the agent’s performance across different data scales.
We retrain Llama3.1-70B on different subsets of the original \dataset, with 10\%, 20\%, 50\%, and 90\% ratios, and compare the performance to that obtained with the full dataset.

\begin{figure}[t!]
  \centering
  \includegraphics[width=0.48\textwidth]{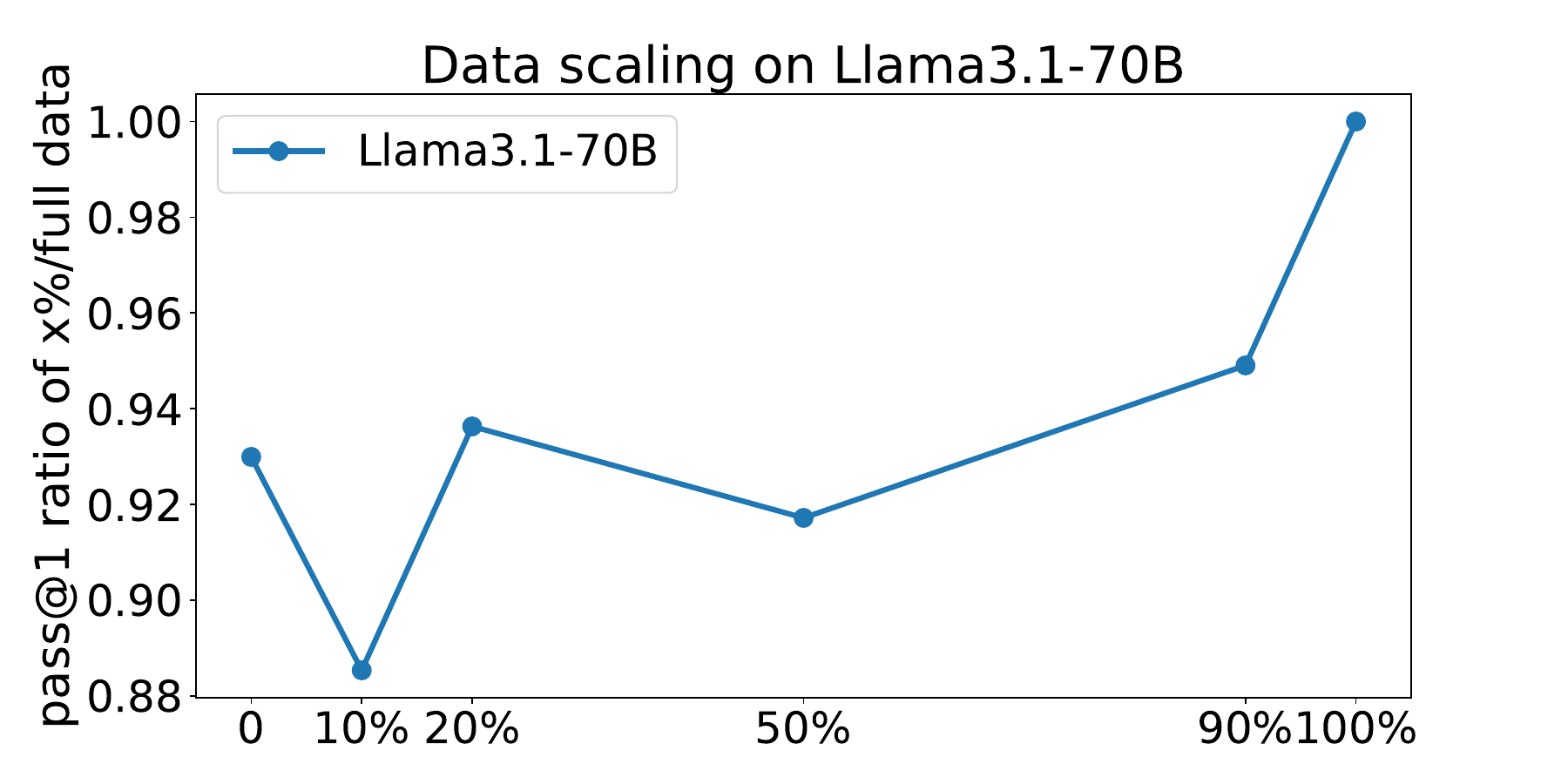}
  \caption{\textbf{Performance over different proportions of training data.}
  The x-axis represents data proportion and the y-axis represents the $\text{pass}@1$ \textit{ratio} between training with a proportion of data and with the full dataset.}
  \label{fig:exp-data-scaling}
\end{figure}

The results are in Figure~\ref{fig:exp-data-scaling}.
The $\text{pass}@1$ fluctuates as the ratio rises to 50\%, at times dropping below that of the model without training.
The performance moderately increases at the ratio of 90\%.
This result suggests that, despite introducing task-relevant information, the limited data size hinders the model’s generalization across scene layouts and may degrade performance.
A critical threshold of training data is required to maximize the agent's ability to generalize and handle the diversity of industrial scene generation.
Our experiment confirms that the amount of training data is vital in achieving high-precision scene generation.

\section{Related Work}
Generating scenes from textual descriptions has broad applicability across various fields such as indoor environments~\citep{chang2017sceneseer, song2023roomdreamer}, road traffic~\citep{kar2019meta, tan2021scenegen}, and city planning~\citep{lipp2011interactive, qin2024urbanevolver}.
A significant body of work focuses on generating structured 3D scenes by arranging modular and interactive objects with LLMs based on textual inputs.
Unlike approaches that directly synthesize 3D representations such as radiance fields and point clouds~\citep{bai2023componerf, cohen2023set, po2024compositional}, these methods utilize pre-defined object assets, enabling the integration of generated scenes into downstream applications like programming in software.
Studies using LLMs to generate 3D scenes introduce various techniques to control scene components.
LayoutGPT~\citep{feng2024layoutgpt}, 3D-GPT~\citep{sun20233d}, and SceneX~\citep{zhou2024scenex} directly generate configurations of the scene, assuming that the LLMs are reliable for accurate arrangement.
Other studies adjust the scene's spatial layout to better adhere to the text.
LLMR~\citep{de2024llmr} relies on validation to refine the LLM's layout planning.
Some methods treat layout information in the description as constraints.
SceneCraft~\citep{hu2024scenecraft} models these constraints using learnable scoring functions and optimizes these scores.
Holodeck~\citep{yang2024holodeck} employs depth-first search on a relational constraint graph to find a placement satisfying the constraints.
GALA3D~\citep{zhou2024gala3d} expresses the constraints with 3D Gaussian splatting representations to refine the layout.
\model differentiates from them by regarding layout information as computable and refinable data structures, enabling precise placement through calculation and reasoning.

\section{Conclusion}
This paper introduces \model, an LLM-based agent that automates the precise generation of industrial scenes, translating descriptions into C\# code using Tecnomatix API.
We propose a layout analysis pipeline within \model that ensures accurate spatial positioning and compliance with complex numerical constraints in industrial layouts.
\model uses a structured, calculable format to represent spatial specifications
and infer overall layout planning through computation, iterative verification, and refinement.
Furthermore, we fine-tune open-source LLMs with our constructed dataset \dataset to improve their performance in \model, approaching the capabilities of API-based models.
We believe that \model can significantly enhance the efficiency and accuracy of industrial manufacturing and provide insights for further advancements in the scene generation field.
\section*{Limitations}
\vpara{Broader applications.}
Our experiments predominantly focus on object assets and textual descriptions within industrial contexts.
However, our proposed method for precise layout generation is not limited to such environments and holds potential for broader application in general scene generation tasks.
Specifically, this method can enhance the quantitative accuracy of generated results across various domains.
We plan to expand this work to more generalized scene generation by incorporating standardized benchmarks and robust evaluation methodologies in the future.

\vpara{Evaluation methods.}
Due to the lack of automated tool suites for industrial software, our evaluation relies on manual evaluation, which is labor-intensive and inherently susceptible to subjective bias.
Future work could explore developing automated evaluation methods to assess the generated scene's compliance with quantitative constraints, particularly in industrial software applications.

\section*{Acknowledgments}
{\small
This work is supported by the Natural Science Foundation of China (NSFC) 62495063, 62276148, and research grants from Siemens and China Datang Corporation Ltd.
The GPU compute used in this work is sponsored by Zhipu AI. 
We would like to thank the members of Siemens Ltd. for their technical support and assistance in the evaluation.
The corresponding author: Yuxiao Dong (yuxiaod@tsinghua.edu.cn). 
}

\bibliography{acl_latex}

\clearpage\newpage
\appendix

\label{sec:appendix}
\section{Appendix}

\subsection{Comparison of \model with Holodeck}

\begin{table*}[t!]
  \centering
  \caption{\textbf{Results of the original \model and \model with Holodeck's placement planning approach.}}
  \begin{tabular}{l|cccccc}
    \toprule
    \multirow{2}{*}{\begin{tabular}[x]{@{}l@{}}\textbf{\model's} \\ \textbf{placement planning approach}\end{tabular}} & \multicolumn{6}{c}{\textbf{$\text{pass}@1$}} \\
    \cmidrule(lr){2-7}
     & Geo. & Pos. & Quant. & Comp. & Fuzz. & Overall \\
    \midrule
    original & 64.0 & 80.0 & 76.7 & 64.0 & 95.0 & 73.0 \\
    from Holodeck & 21.3 & 11.1 & 23.3 & 24.0 & 90.0 & 28.0 \\
    \bottomrule
  \end{tabular}
  \label{tab:exp-holodeck}
\end{table*}

To facilitate a comparative analysis with \model, we select Holodeck~\cite{yang2024holodeck}, a recently developed scene generation framework with publicly available source code.
Specifically, we integrate Holodeck's placement planning module into the \model's framework by replacing the layout extraction and placement assignment components of \model with the corresponding modules from Holodeck.
We directly utilize Holodeck's placement planning code, making minimal modifications to the scene configurations and examples in the prompts to align with the requirements of industrial scene generation tasks, while leaving the remainder of the code unchanged.
To evaluate performance, we compare the original \model and the modified version incorporating Holodeck's placement planning, employing Llama3.1-70B without fine-tuning as the backbone.

The $\text{pass}@1$ results for both variants are summarized in Table~\ref{tab:exp-holodeck}.
The results indicate that the original \model substantially outperforms the version employing Holodeck's placement planning module.
The substantial performance gap across all categories, particularly in geometric arrangement and positional details, highlights Holodeck's limitations in addressing the quantitative spatial constraints critical to industrial scene generation.
These results underscore the effectiveness of \model's layout analysis pipeline, demonstrating that previous scene generation methods, while effective in general domains, are inadequate for the specialized demands of industrial scene generation.

\subsection{Datasheet}
\label{app:datasheet}
We provide the datasheet of \dataset following the recommendations of~\citet{gebru2021datasheets}.

\subsubsection{Motivation}
\textbf{For what purpose was the dataset created?}
The dataset was created for fine-tuning language models to enhance their performance in industrial scene generation.

\textbf{Who created the dataset (e.g., which team, research group) and on behalf of which entity (e.g., company, institution, organization)?} The dataset was created by the authors of this paper.

\subsubsection{Compositions}
\textbf{What do the instances that comprise the dataset represent (e.g., documents, photos, people, countries)?} The instances that comprise the dataset represent the action trajectories of SceneGenAgent, a language model-based agent. The actions are of three types: assign placement, verify placement, and reassign placement.

\textbf{How many instances are there in total (of each type, if appropriate)?} There are 21756 instances in total. 20688 instances are for training, including 5704 for assigning placements, 12397 for verifying placements, and 2587 for reassigning placements. 1068 instances are for validation, including 300 for assigning placements, 636 for verifying placements, and 132 for reassigning placements.

\textbf{Does the dataset contain all possible instances or is it a sample (not necessarily random) of instances from a larger set?} The dataset is synthesized using language models and contains all possible instances we can sample.

\textbf{What data does each instance consist of?} Each instance consists of the input and output of a language model on assigning, verifying, and reassigning placement tasks, as described in Section~\ref{sec:dataset-construction}.

\textbf{Is there a label or target associated with each instance?} Yes, every instance contains a model output as the target.

\textbf{Is any information missing from individual instances?} No.

\textbf{Are there recommended data splits (e.g., training, development/validation, testing)?} Yes, as described in Section~\ref{sec:dataset-construction}.

\textbf{Are there any errors, sources of noise, or redundancies in the dataset?} Potential errors may be introduced by the language models used for data synthesis during reasoning and calculation.

\textbf{Is the dataset self-contained, or does it link to or otherwise rely on external resources (e.g., websites, tweets, other datasets)?} The dataset is self-contained.

\textbf{Does the dataset contain data that might be considered confidential (e.g., data that is protected by legal privilege or by doctor-patient confidentiality, data that includes the content of individuals’ non-public communications)?} No.

\textbf{Does the dataset contain data that, if viewed directly, might be offensive, insulting, threatening, or might otherwise cause anxiety?} No.

\subsubsection{Collection Process}
\textbf{How was the data associated with each instance acquired?} The descriptions as the agent's input are synthesized by iteratively rewriting 20 hand-written descriptions with Llama3.1-70B. The inputs and outputs of the instances are acquired by feeding the synthesized descriptions into the agent and collecting the inputs and outputs from multiple stages of the agent.

\textbf{What mechanisms or procedures were used to collect the data (e.g., hardware apparatuses or sensors, manual human curation, software programs, software APIs)?} The data is collected using Llama3.1-70B, GPT-4 model API, and Llama3.1-8B for generation.

\textbf{Over what timeframe was the data collected?} August 2024.

\textbf{Were any ethical review processes conducted (e.g., by an institutional review board)?} No.

\subsubsection{Preprocessing/cleaning/labeling}
\textbf{Was any preprocessing/cleaning/labeling of the data done (e.g., discretization or bucketing, tokenization, part-of-speech tagging, SIFT feature extraction, removal of instances, processing of missing values)?} Yes, as described in Section~\ref{sec:dataset-construction} of this paper.

\textbf{Was the "raw" data saved in addition to the preprocessed/cleaned/labeled data?} No.

\textbf{Is the software that was used to preprocess/clean/label the data available?} Yes.

\subsubsection{Uses}
\textbf{Has the dataset been used for any tasks already?} Yes, as described in Section~\ref{sec:data-tuning} of this paper.

\textbf{Is there a repository that links to any or all papers or systems that use the dataset?} Yes.

\textbf{What (other) tasks could the dataset be used for?} N/A.

\textbf{Is there anything about the composition of the dataset or the way it was collected and preprocessed/cleaned/labeled that might impact future uses?} No.

\subsubsection{Distribution}
\textbf{Will the dataset be distributed to third parties outside of the entity (e.g., company, institution, organization) on behalf of which the dataset was created?} Yes, the dataset is publicly distributed.

\textbf{How will the dataset be distributed (e.g., tarball on website, API, GitHub)?} The dataset is publicly available on GitHub.

\textbf{When will the dataset be distributed?} The dataset has been distributed since October 2024.

\textbf{Will the dataset be distributed under a copyright or other intellectual property (IP) license, and/or under applicable terms of use (ToU)?} The dataset is distributed under an Apache-2.0 license.

\textbf{Have any third parties imposed IP-based or other restrictions on the data associated with the instances?} No.

\textbf{Do any export controls or other regulatory restrictions apply to the dataset or to individual instances?} No.

\subsubsection{Maintenance}
\textbf{Who will be supporting/hosting/maintaining the dataset?} The authors of this paper.

\textbf{How can the owner/curator/manager of the dataset be contacted (e.g., email address)?} Via the authors' email addresses or GiHub Issues.

\textbf{Is there an erratum?} Not at this time.

\textbf{Will the dataset be updated (e.g., to correct labeling errors, add new instances, delete instances)?} Yes, when labeling errors are found in the dataset.

\textbf{Will older versions of the dataset continue to be supported/hosted/maintained?} Yes.

\textbf{If others want to extend/augment/build on/contribute to the dataset, is there a mechanism for them to do so?} Yes, by creating pull requests on GitHub.

\subsection{\dataset Data Examples}
\label{app:data-examples}
\vpara{Assign placement}

\noindent Input:

\begin{tcolorbox}[left=0mm,right=0mm,top=0mm,bottom=0mm,boxsep=1mm,arc=0mm,boxrule=0pt, frame empty, breakable]
\small
\begin{lstlisting}
You are given a description of a workstation wherein a series of objects exist, with their respective positions mentioned in the form of coordinates and relative positioning to one another. The description is as follows:
```
One Welding Table for spot welding, with a Turntable at [1500, 2500, 0], is equipped with an ABB Robot IRB6600, positioned 2.6 meters to the right and 2.5 meters back from the Turntable.
```
The objects mentioned in the description are: ["Welding Table", "Turntable", "ABB Robot IRB6600"]
Their known positions are as follows:
```
[
    {
        "name": "Turntable",
        "position": "[1500, 2500, 0]",
        "orientation": "0"
    }
]
```
Their relative positions are as follows:
```
[
    {
        "object 1": "ABB Robot IRB6600",
        "relation": "2.6 meters to the right and 2.5 meters back",
        "object 2": "Turntable"
    }
]
```
For directions and orientations, consider the front as the positive direction of the x-axis, the back as the negative direction of the x-axis, the left as the positive direction of the y-axis, and the right as the negative direction of the y-axis.
The orientation of an object is what direction the object faces, including an angle rotated counterclockwise on the ground in degrees or towards other objects. For example, rotating for 0 degrees is to face the front, 90 degrees is to face the left, 180 degrees is to face the back, and 270 degrees is to face the right.
Based on the description and position information, you should assign coordinates for every objects. The coordinates are three values in millimeters in the form of "[x, y, 0]" with no unit after the brackets, with x and y representing the components of the coordinate on the x-axis and y-axis respectively.
You should do the following:
1. Rewrite every relative position into the increment of one object's coordinate relative to another object's coordinate.
Write the increment in specific values [+-x, +-y, 0] instead of placeholders. One relative position should only be listed once.
If you need to set distances or directions yourself, the Euclidean distances between objects should generally be greater than 1 meter, which is 1000 in coordinates, unless stated otherwise in the description.
2. For each relative position, if one object has a declared coordinate, calculate the coordinate of the other object. Then, list the positional information of each object, including coordinates and orientations.
You should first perform calculations to get the coordinates in the analysis part, then come up with the final positions.
3. For the remaining objects without coordinates, assign valid coordinates for them based on the restrictions of object position, direction, orientation, and relation.
The coordinate values should generally be from [-5000, 0] or [0, 5000] intervals unless stated otherwise in the description.
The Euclidean distances between objects should generally be greater than 1 meter, which is 1000 in coordinates, unless stated otherwise in the description.
All objects are on the ground with no vertical relative positions.
To avoid overlapping, coordinates assigned in this step should be different from existing objects and each other, with the exception of Guarding which surrounds the objects and whose position represents its center which is not occupied.
You should first perform calculations to get the coordinates and deduce whether the coordinates of all objects meet the above conditions in the analysis part, then come up with the final positions.
After this step, every object in the description should have a coordinate.
You should list the positional information, including coordinates and orientations, of all objects in Step 3.
You must not skip any step.
You should return in the following format:
```
#Step 1: Rewrite Relative Position#
Analysis: <analysis>
New Relative Positions:
[
    {
        "object 1": "<object name 1>",
        "relation": "[+-x, +-y, 0]",
        "object 2": "<object name 2>"
    },
    ...
]

#Step 2: Calculate Coordinates#
Analysis: <analysis and calculation>
Positions:
[
    {
        "name": "<object name>",
        "position": "[x, y, 0]",
        "orientation": "<orientation>"
    },
    ...
]

#Step 3: Assign Positions#
Analysis: <analysis and calculation>
Positions:
[
    {
        "name": "<object name>",
        "position": "[x, y, 0]",
        "orientation": "<orientation>"
    },
    ...
]
```
Do not say anything else.
\end{lstlisting}
\end{tcolorbox}

\noindent Output:

\begin{tcolorbox}[left=0mm,right=0mm,top=0mm,bottom=0mm,boxsep=1mm,arc=0mm,boxrule=0pt, frame empty, breakable]
\small
\begin{lstlisting}
#Step 1: Rewrite Relative Position#
Analysis: Given that the direction towards the right is considered negative on the y-axis and the direction back is considered negative on the x-axis, the ABB Robot IRB6600's position relative to the Turntable can be precisely defined in terms of coordinate changes. As it is 2.6 meters to the right, this translates to -2600 on the y-axis, and being 2.5 meters back translates to -2500 on the x-axis, considering the origin at the Turntable's position [1500, 2500, 0].
New Relative Positions:
[
    {
        "object 1": "ABB Robot IRB6600",
        "relation": "[-2500, -2600, 0]",
        "object 2": "Turntable"
    }
]

#Step 2: Calculate Coordinates#
Analysis: Given that the Turntable is located at [1500, 2500, 0], applying the relative position to find the ABB Robot IRB6600's position would involve subtracting 2500 from the x coordinate and 2600 from the y coordinate of the Turntable's position.
Positions:
[
    {
        "name": "Turntable",
        "position": "[1500, 2500, 0]",
        "orientation": "0"
    },
    {
        "name": "ABB Robot IRB6600",
        "position": "[-1000, -100, 0]",
        "orientation": "0"
    }
]

#Step 3: Assign Positions#
Analysis: With both the Turntable and the ABB Robot IRB6600's positions now known, the Welding Table's position needs to be set while ensuring it respects the minimum distance criterion from other objects. Given its absence in the relative position description, a new position must be chosen. Considering the spatial layout, placing it at a considerable distance to avoid collision and maintain operational efficiency is optimal. A valid choice is ensuring it's more than 1000 mm (1 meter) away from both the Turntable and the ABB Robot.
Positions:
[
    {
        "name": "Turntable",
        "position": "[1500, 2500, 0]",
        "orientation": "0"
    },
    {
        "name": "ABB Robot IRB6600",
        "position": "[-1000, -100, 0]",
        "orientation": "0"
    },
    {
        "name": "Welding Table",
        "position": "[0, 4500, 0]",
        "orientation": "0"
    }
]
\end{lstlisting}
\end{tcolorbox}

\vpara{Check placement}

\noindent Input:

\begin{tcolorbox}[left=0mm,right=0mm,top=0mm,bottom=0mm,boxsep=1mm,arc=0mm,boxrule=0pt, frame empty, breakable]
\small
\begin{lstlisting}
You are given a description of a workstation, which is used to build a scene in Process Simulate (PS).
The description includes several objects and their positions. The description is as follows:
```
One Welding Table for spot welding, with a Turntable at [1500, 2500, 0], is equipped with an ABB Robot IRB6600, positioned 2.6 meters to the right and 2.5 meters back from the Turntable.
```
For directions and orientations, consider the front as the positive direction of the x-axis, the back as the negative direction of the x-axis, the left as the positive direction of the y-axis, and the right as the negative direction of the y-axis.
The orientation of an object is what direction the object faces, including an angle rotated counterclockwise on the ground in degrees or towards other objects. For example, rotating for 0 degrees is to face the front, 90 degrees is to face the left, 180 degrees is to face the back, and 270 degrees is to face the right. The default orientation is to rotate for 0 degrees which makes the object face the positive direction of the x-axis.
Based on the description, we allocate positions for the objects. The allocated positions of objects are as follows:
```
[
  {
    "name": "Turntable",
    "position": "[1500, 2500, 0]",
    "orientation": "0"
  },
  {
    "name": "Welding Table",
    "position": "[-5000, 0, 0]",
    "orientation": "0"
  },
  {
    "name": "ABB Robot IRB6600",
    "position": "[-1900, 0, 0]",
    "orientation": "0"
  }
]
```
However, the positions may represent certain errors which need to be identified.
You should carry out the following actions to check whether there are conflicts in object positions:
1. Calculate each object's positional arrangement. In case objects' positions are described relatively, deduce the corresponding coordinates. All objects exist in a 2D plane with z-coordinate being 0.
2. Identify any existing errors among the positions of the objects:
   - Constraints: The description may provide additional positional constraints which the objects should not violate. The front is the positive direction of the x-axis and the left is the positive direction of the y-axis.
   - Conflicts: The position description may contain inconsistencies among the objects. The position of an object may be calculated in many ways, and if the results obtained from different calculations are not consistent, there are conflicts in the description.
   - Overlap: The Euclidean distances between objects should be greater than 1 meter, which is 1000 in coordinates, or the objects will overlap with each other. Note that one exception is the Guarding, which surrounds the objects inside it and whose position represents its center which is not occupied. Also note that objects can overlap with each other if the description requires them to do so, e.g. their coordinates are explicitly specified.
3. Report whether there are positional errors. If you cannot deduce whether there are any errors - for example, when precise positions cannot be computed, please report that there are no errors.
You should return in the following format:
```
Relations: <Calculate the coordinates and positional relations>
Analysis: <Based on the calculated positions, deduce whether there are errors>
Error: <Based on the analysis, decide whether there are errors in the description. Choose between "Yes" and "No". If you can not determine, choose "No">
```
Do not say anything else.
\end{lstlisting}
\end{tcolorbox}

\noindent Output:

\begin{tcolorbox}[left=0mm,right=0mm,top=0mm,bottom=0mm,boxsep=1mm,arc=0mm,boxrule=0pt, frame empty, breakable]
\small
\begin{lstlisting}
Relations: The Turntable is at [1500, 2500, 0]. The ABB Robot IRB6600 is positioned 2.6 meters to the right (-2600 on y-axis) and 2.5 meters back (-2500 on x-axis) from the Turntable, so it should be at [-1000, -100, 0].

Analysis: Based on the calculations, there's a mistake in the positions. The positions in the provided description do not match the calculated position of the ABB Robot IRB6600 relative to the Turntable. Also, the Welding Table's position seems unrelated to the provided scenario description and doesn't match any calculated relation from the given information.

Error: Yes
\end{lstlisting}
\end{tcolorbox}

\vpara{Reassign placement}

\noindent 1st round input:

\begin{tcolorbox}[left=0mm,right=0mm,top=0mm,bottom=0mm,boxsep=1mm,arc=0mm,boxrule=0pt, frame empty, breakable]
\small
\begin{lstlisting}
You are given a description of a workstation wherein a series of objects exist, with their respective positions mentioned in the form of coordinates and relative positioning to one another. The description is as follows:
```
One Welding Table for spot welding, with a Turntable at [1500, 2500, 0], is equipped with an ABB Robot IRB6600, positioned 2.6 meters to the right and 2.5 meters back from the Turntable.
```
The objects mentioned in the description are: ["Welding Table", "Turntable", "ABB Robot IRB6600"]
Their known positions are as follows:
```
[
    {
        "name": "Turntable",
        "position": "[1500, 2500, 0]",
        "orientation": "0"
    }
]
```
Their relative positions are as follows:
```
[
    {
        "object 1": "ABB Robot IRB6600",
        "relation": "2.6 meters to the right and 2.5 meters back",
        "object 2": "Turntable"
    }
]
```
For directions and orientations, consider the front as the positive direction of the x-axis, the back as the negative direction of the x-axis, the left as the positive direction of the y-axis, and the right as the negative direction of the y-axis.
The orientation of an object is what direction the object faces, including an angle rotated counterclockwise on the ground in degrees or towards other objects. For example, rotating for 0 degrees is to face the front, 90 degrees is to face the left, 180 degrees is to face the back, and 270 degrees is to face the right.
Based on the description and position information, you should assign coordinates for every objects. The coordinates are three values in millimeters in the form of "[x, y, 0]" with no unit after the brackets, with x and y representing the components of the coordinate on the x-axis and y-axis respectively.
You should do the following:
1. Rewrite every relative position into the increment of one object's coordinate relative to another object's coordinate.
Write the increment in specific values [+-x, +-y, 0] instead of placeholders. One relative position should only be listed once.
If you need to set distances or directions yourself, the Euclidean distances between objects should generally be greater than 1 meter, which is 1000 in coordinates, unless stated otherwise in the description.
2. For each relative position, if one object has a declared coordinate, calculate the coordinate of the other object. Then, list the positional information of each object, including coordinates and orientations.
You should first perform calculations to get the coordinates in the analysis part, then come up with the final positions.
3. For the remaining objects without coordinates, assign valid coordinates for them based on the restrictions of object position, direction, orientation, and relation.
The coordinate values should generally be from [-5000, 0] or [0, 5000] intervals unless stated otherwise in the description.
The Euclidean distances between objects should generally be greater than 1 meter, which is 1000 in coordinates, unless stated otherwise in the description.
All objects are on the ground with no vertical relative positions.
To avoid overlapping, coordinates assigned in this step should be different from existing objects and each other, with the exception of Guarding which surrounds the objects and whose position represents its center which is not occupied.
You should first perform calculations to get the coordinates and deduce whether the coordinates of all objects meet the above conditions in the analysis part, then come up with the final positions.
After this step, every object in the description should have a coordinate.
You should list the positional information, including coordinates and orientations, of all objects in Step 3.
You must not skip any step.
You should return in the following format:
```
#Step 1: Rewrite Relative Position#
Analysis: <analysis>
New Relative Positions:
[
    {
        "object 1": "<object name 1>",
        "relation": "[+-x, +-y, 0]",
        "object 2": "<object name 2>"
    },
    ...
]

#Step 2: Calculate Coordinates#
Analysis: <analysis and calculation>
Positions:
[
    {
        "name": "<object name>",
        "position": "[x, y, 0]",
        "orientation": "<orientation>"
    },
    ...
]

#Step 3: Assign Positions#
Analysis: <analysis and calculation>
Positions:
[
    {
        "name": "<object name>",
        "position": "[x, y, 0]",
        "orientation": "<orientation>"
    },
    ...
]
```
Do not say anything else.
\end{lstlisting}
\end{tcolorbox}

\noindent 1st round output (do not calculate losses for training):

\begin{tcolorbox}[left=0mm,right=0mm,top=0mm,bottom=0mm,boxsep=1mm,arc=0mm,boxrule=0pt, frame empty, breakable]
\small
\begin{lstlisting}
[
  {
    "name": "Turntable",
    "position": "[1500, 2500, 0]",
    "orientation": "0"
  },
  {
    "name": "ABB Robot IRB6600",
    "position": "[-1100, 0, 0]",
    "orientation": "0"
  },
  {
    "name": "Welding Table",
    "position": "[0, 3500, 0]",
    "orientation": "0"
  }
]
\end{lstlisting}
\end{tcolorbox}

\noindent 2nd round input:

\begin{tcolorbox}[left=0mm,right=0mm,top=0mm,bottom=0mm,boxsep=1mm,arc=0mm,boxrule=0pt, frame empty, breakable]
\small
\begin{lstlisting}
Your position allocation contains the following error:
```
Relations: The Turntable is at [1500, 2500, 0]. The ABB Robot IRB6600 is positioned 2.6 meters to the right (negative direction of the y-axis) and 2.5 meters back (negative direction of the x-axis) from the Turntable, resulting in its position being [1200, 2500-2600, 0] = [1200, -100, 0]. The Welding Table location is not described relative to the Turntable or the ABB Robot IRB6600, and its given position seems to be an absolute one without relational context from the description provided.

Analysis: Given the positions based on the description:
- Turntable at [1500, 2500, 0]
- ABB Robot IRB6600 calculated at [1200, -100, 0] which contradicts the given "[ -1100, 0, 0]"
- Welding Table given at "[0, 3500, 0]" without relative positioning stated in the description.

There seem to be inconsistencies in the provided positions when compared with calculations based on the provided descriptions, particularly for the ABB Robot IRB6600. Additionally, without explicit relational descriptions for the Welding Table, its position remains arbitrary without conflict but lacks verification.
```
Please write the response again, fixing the errors.
Only generate the response. Do not say anything else.
\end{lstlisting}
\end{tcolorbox}

\noindent 2nd round output (calculate losses for training):

\begin{tcolorbox}[left=0mm,right=0mm,top=0mm,bottom=0mm,boxsep=1mm,arc=0mm,boxrule=0pt, frame empty, breakable]
\small
\begin{lstlisting}
#Step 1: Rewrite Relative Position#
Analysis: Given that the direction towards the right is considered negative on the y-axis and the direction back is considered negative on the x-axis, the ABB Robot IRB6600's position relative to the Turntable can be precisely defined in terms of coordinate changes. As it is 2.6 meters to the right, this translates to -2600 on the y-axis, and being 2.5 meters back translates to -2500 on the x-axis, considering the origin at the Turntable's position [1500, 2500, 0].
New Relative Positions:
[
    {
        "object 1": "ABB Robot IRB6600",
        "relation": "[-2500, -2600, 0]",
        "object 2": "Turntable"
    }
]

#Step 2: Calculate Coordinates#
Analysis: Given that the Turntable is located at [1500, 2500, 0], applying the relative position to find the ABB Robot IRB6600's position would involve subtracting 2500 from the x coordinate and 2600 from the y coordinate of the Turntable's position.
Positions:
[
    {
        "name": "Turntable",
        "position": "[1500, 2500, 0]",
        "orientation": "0"
    },
    {
        "name": "ABB Robot IRB6600",
        "position": "[-1000, -100, 0]",
        "orientation": "0"
    }
]

#Step 3: Assign Positions#
Analysis: With both the Turntable and the ABB Robot IRB6600's positions now known, the Welding Table's position needs to be set while ensuring it respects the minimum distance criterion from other objects. Given its absence in the relative position description, a new position must be chosen. Considering the spatial layout, placing it at a considerable distance to avoid collision and maintain operational efficiency is optimal. A valid choice is ensuring it's more than 1000 mm (1 meter) away from both the Turntable and the ABB Robot.
Positions:
[
    {
        "name": "Turntable",
        "position": "[1500, 2500, 0]",
        "orientation": "0"
    },
    {
        "name": "ABB Robot IRB6600",
        "position": "[-1000, -100, 0]",
        "orientation": "0"
    },
    {
        "name": "Welding Table",
        "position": "[0, 4500, 0]",
        "orientation": "0"
    }
]
\end{lstlisting}
\end{tcolorbox}

\subsection{Details on \dataset Construction}
\subsubsection{Instruction Construction}
\label{app:instruct-construct-detail}
During construction of descriptions as instructions in \dataset, we devise the following rewriting methods:
\begin{itemize}[leftmargin=*,itemsep=0pt,parsep=0.5em,topsep=0.3em,partopsep=0.3em]
    \item \textbf{Object addition/replacement}: Introduce new objects or substitute one or more objects in the description with 2 to 4 objects randomly picked from the object library.
    \item \textbf{Location specification}: Specify the locations of the objects in the description with reasonable coordinate values. We advise the LLM to choose coordinate values greater than -5000 and less than 5000. The distance between objects should be greater than 1 meter.
    \item \textbf{Relation specification}: Specify the relations between two objects mentioned in the description. The relation may include distance, direction, or orientation. We advise the LLM that the distance between objects should be greater than 1 meter and less than 5 meters.
    \item \textbf{Quantity modification}: Adjust the quantity of an object to a value between 3 and 10 if it is not assigned a placement or relations.
    \item \textbf{Fuzzy expressions}: Replace precise numerical relations between two objects with ambiguous or qualitative expressions.
    This is to get more data where the agent must set values actively, encouraging the agent to infer values independently.
    \item \textbf{Rephrasing}: Restate the description using alternative wording while retaining its meaning.
\end{itemize}

During each rewriting iteration, a rewriting method is selected probabilistically from a multinomial distribution.
The sampling probability of each method is $(\frac{5}{24}, \frac{6}{24}, \frac{6}{24}, \frac{1}{24}, \frac{5}{24}, \frac{1}{24})$, encouraging modifications of objects and layout in the description.
Descriptions are sampled from the description pool with equal probability.

\subsubsection{Trajectory Collection}
\label{app:trajectory-collection-detail}
We collect the trajectories of \model using the constructed instruction dataset.
However, collecting trajectories from all parts of \model presents significant challenges, particularly because the object retrieval part must be able to process highly arbitrary descriptions, which are difficult for the Evol-Instruct method we adopted to produce ~\citep{zeng2024automatic}.
To address this, we select the parts of the agent for data collection by observing Llama3.1-70B's performance in each part.
Consequently, we select the placement assignment stage for trajectory collection.

\begin{table*}[t!]
    \centering
    \caption{Hyper-parameters for model training with \dataset.}
    \begin{tabular}{lccccc}
\toprule 
\textbf{Model} & \textbf{Peak Learning Rate} & \textbf{Batch Size} & \textbf{LoRA Rank} & \textbf{LoRA alpha} & \textbf{\#Epochs} \\
\midrule 
Qwen2.5-Coder-7B & 1e-5 & 64 & 64 & 128 & 2 \\
Llama3.1-8B & 1e-5 & 256 & 64 & 128 & 2 \\
GLM4-9B & 1e-5 & 64 & 64 & 128 & 8 \\
Codestral-22B & 1e-4 & 256 & 64 & 128 & 8 \\
CodeLlama-34B & 1e-4 & 256 & 64 & 128 & 8 \\
Llama3.1-70B & 1e-4 & 256 & 64 & 128 & 8 \\
\bottomrule
    \end{tabular}
    \label{tab:hyper-parameters}
\end{table*}

\subsection{Details on Industrial Scene Generation Settings}
\label{app:detail-setting}
\subsubsection{Object Library}
\label{app:detail-object-library}
The object library we use is from the Siemens Tecnomatix Process Simulate tutorial, containing 20 different models of 10 types, including equipment and facilities like robotic arms, work tables, conveyor belts, cabinets, and guardings, for modeling and simulating workstations with different layouts.

\subsubsection{Evaluation Settings}
\label{app:detail-evaluation-settings}
\vpara{Evaluation Method.}
We manually inspect the scenes and compare them against their descriptions to determine their quality during evaluation.
Due to the absence of testing tools and the non-uniqueness of correct generation in scene generation tasks, we can only measure the generation quality by visually inspecting the scenes.
Previous studies on scene generation~\cite{yang2024holodeck, hu2024scenecraft, zhou2024gala3d, zhou2024scenex} utilize text-to-image CLIP scores~\cite{radford2021learning} to assess scene quality automatically.
However, this metric is insufficient to evaluate a scene's quantitative accuracy, which is a key focus of our assessment.
Although we explored the use of multimodal LLMs, such as GPT-4o, to verify whether scene images align with the quantitative requirements in the descriptions, adding grid lines in the scenes to aid in distance identification, current models struggle to extract precise quantitative information from images, making them unsuitable for evaluation.
We anticipate that advancements in testing tools and multimodal LLMs may enable automated evaluation.

\vpara{Collision Criterion.}
We require the Euclidean distance between two objects at least 1 meter to avoid collision. 
This criterion is introduced by engineers with expertise in industrial scene modeling and incorporated into our work to align with established real-world practices, adequate for preventing collisions.
If additional factors like volume and projection area were considered, they would serve as supplementary distance constraints. In this case, the agent would adhere to these constraints, analogous to the one-meter requirement in the placement assignment stage. Specifically, this would involve converting relations to coordinate differences, allocating placement to the objects, and verifying the placement to ensure the scene's coherence.

\vpara{Evaluation of Open-Source LLMs.}
We evaluate open-source LLMs by applying them during the placement assignment stage with Llama3.1-70B managing the remaining operations.
This setting is due to the way we construct the \dataset dataset.
We refer to Llama3.1-70B's capabilities to select the parts of the agent for data collection, which is the placement assignment stage, as discussed in Section~\ref{sec:dataset-construction} and Appendix~\ref{app:trajectory-collection-detail}.
However, smaller models struggle in operations beyond placement assignment, making it unable to evaluate their performance.
To address this, we use Llama3.1-70B to handle operations other than placement assignment for all open models to enable the evaluation of smaller models and provide evidence that fine-tuning with \dataset improves the performance of these models in \model.

\subsection{Training Hyper-Parameters}
\label{app:train-hyperparams}

For models fine-tuned with \dataset for the results in Table~\ref{tab:exp-main}, their training hyper-parameters are provided in Table~\ref{tab:hyper-parameters}.

\subsection{Prompts}
\label{app:prompts}
\subsubsection{\model}
\label{app:agent-prompts}
\vpara{Object Retrieval}
\begin{tcolorbox}[left=0mm,right=0mm,top=0mm,bottom=0mm,boxsep=1mm,arc=0mm,boxrule=0pt, frame empty, breakable]
\small
\begin{lstlisting}
You are given a description of a workstation, which is used to build a scene in Process Simulate (PS).
The original description is as follows:
```
{{ prompt }}
```
You should follow the steps below:
1. First, find all objects mentioned in the description and should appear in the workstation according to the description. Objects with or without specific positions should all be listed. If the description mentions multiple instances of the same object, list the object repeatedly for each instance. For example, if "ten tables" are mentioned, then include "table" ten times. Note that a "station", a "workstation", or a "scene" is not an object.
2. Then, fix the names of objects that are not correct. The objects that may appear in the description are from the permission list: [Kuka Robot KR125, Kuka Robot KR350, ABB Robot IRB6600, YASKAWA Robot ma01800, Welding Table, Turntable, Cabinet, ValveStand, Conveyor, Guarding]. For every object you find, change it so that it meets all the requirements:
   - All objects found in the previous step is independent of each other. Do not merge them into one object.
   - If the object is not from the permission list, including upper/lower cases, replace it with an object in the permission list that satisfies the restrictions in the description.
   - If there are objects with general types, replace them with specific objects from the list.
   - The object type must satisfy the restrictions in the description. For example, the description may state that some objects must be the same/different.
If the description requires you to add objects or there are no objects in the scene, add some objects that meet the description restrictions.
Only fix the objects that are already found and do not add additional objects.
Return all objects after correction, including ones that are correct and ones that are changed to be correct.
3. Finally, you should modify the original description:
   - Rewrite the description like an industrial modeling engineer would in accurate wordings.
   - Include all objects found in the last step, including upper/lower cases.
   - All objects found in the previous step is independent of each other. Do not merge them into one object.
   - Keep the positional, directional, and orientational information of objects.
   - Remove the color, shape, length, height, and function attributes of objects.
Return the rewritten description with one single sentence in one line and keep all positional information.

You should return your analysis, objects and description in the format below:
```
#Step 1: Find all objects#
Analysis: <analysis>
Objects: ["ObjA", "ObjB", ...]

#Step 2: Fix object names#
Analysis: <analysis>
Objects: ["ObjC", "ObjD", ...]

#Step 3: Rewrite description#
Analysis: <analysis>
New Description: <new description>
```
Do not say anything else.
\end{lstlisting}
\end{tcolorbox}

\vpara{Layout Extraction}
\begin{tcolorbox}[left=0mm,right=0mm,top=0mm,bottom=0mm,boxsep=1mm,arc=0mm,boxrule=0pt, frame empty, breakable]
\small
\begin{lstlisting}
You are given a description of a workstation, which is used to build a scene in Process Simulate (PS).
The description includes several objects and their positions. The description is as follows:
```
{{ prompt }}
```
The objects mentioned in the description are: {objects}
You should perform the following steps:
1. First, based on the provided object list, list all objects that are mentioned in the description and should appear in the scene according to the description. Objects with or without specific positions should all be listed.
If the description mentions multiple instances of the same object, list the object name repeatedly for each instance and assign numbers in the object names to tell them apart.
You should return a list of objects in the following format:
```
Objects:
["ObjA", "ObjB 1", "ObjB 2", ...]
```
In the example, "ObjB 1" and "Obj 2" are two objects of the same kind. You should assign numbers to tell them apart.
You should not return unspecified object names. Instead, infer them to object names in the list.

2. Then, find all positional information of objects and reference points directly provided in the description. The positional information of objects include coordinates and orientations.
The coordinates are in the form of [x, y, 0], which should contain three values in millimeters in the form of "[x, y, 0]" with no unit after the brackets.
The coordinates of different objects must be different to avoid overlapping with Guarding as an exception, which surrounds the objects inside it and whose position represents its center which is not occupied.
For directions and orientations, consider the front as the positive direction of the x-axis, the back as the negative direction of the x-axis, the left as the positive direction of the y-axis, and the right as the negative direction of the y-axis.
The orientation of an object is what direction the object faces, including an angle rotated counterclockwise on the ground in degrees or towards other objects. For example, rotating for 0 degrees is to face the front, 90 degrees is to face the left, 180 degrees is to face the back, and 270 degrees is to face the right. The default orientation is to rotate for 0 degrees which makes the object face the positive direction of the x-axis.
Only list coordinates directly provided by the description. Do not calculate coordinates by yourself.
You should return a list of positions that appear in the description:
```json
Positions:
[
    {
        "name": "<name>",
        "position": "<absolute position, including coordinate [x, y, 0]>",
        "orientation": "<orientation>"
    },
    ...
]
```
Use JSON format. <name> should either be from the object list or be an reference point.
If no coordinates are provided, return an empty list.

3. Find every relative positions that are mentioned in the description between objects or reference points. The relative positions may include distances, directions, and orientations, and should include specific distances, directions, and orientations instead of ambiguous descriptions if they are provided. The distances between objects or reference points should be outside brackets and in meters. Each pair of objects or reference points should be listed at most once.
You should return a list of relative positions that appear in the description.
```json
Relative Positions:
[
    {
        "object 1": "<name 1>",
        "relation": "<object 1 relative to object 2>",
        "object 2": "<name 2>"
    },
    ...
]
```
Use JSON format. Both <name 1> and <name 2> should be either from the object list or an reference point.
If no relative positions are found, return an empty list.

You should first write your analysis of the description, then return the position information, in the following format:
```
#Step 1: Identify Objects#
Analysis: <analysis>
Objects:
[...]

#Step 2: Absolute Positions#
Analysis: <analysis>
Positions:
[
    ...
]

#Step 3: Relative Positions#
Analysis: <analysis>
Relative Positions:
[
    ...
]
```
Do not say anything else.
\end{lstlisting}
\end{tcolorbox}

\vpara{Placement Assignment}
\begin{tcolorbox}[left=0mm,right=0mm,top=0mm,bottom=0mm,boxsep=1mm,arc=0mm,boxrule=0pt, frame empty, breakable]
\small
\begin{lstlisting}
You are given a description of a workstation wherein a series of objects exist, with their respective positions mentioned in the form of coordinates and relative positioning to one another. The description is as follows:
```
{{ prompt }}
```
The objects mentioned in the description are: {{ objects }}
Their known positions are as follows:
```
{{ positions }}
```
Their relative positions are as follows:
```
{{ relations }}
```
For directions and orientations, consider the front as the positive direction of the x-axis, the back as the negative direction of the x-axis, the left as the positive direction of the y-axis, and the right as the negative direction of the y-axis.
The orientation of an object is what direction the object faces, including an angle rotated counterclockwise on the ground in degrees or towards other objects. For example, rotating for 0 degrees is to face the front, 90 degrees is to face the left, 180 degrees is to face the back, and 270 degrees is to face the right.
Based on the description and position information, you should assign coordinates for every objects. The coordinates are three values in millimeters in the form of "[x, y, 0]" with no unit after the brackets, with x and y representing the components of the coordinate on the x-axis and y-axis respectively.
You should do the following:
1. Rewrite every relative position into the increment of one object's coordinate relative to another object's coordinate.
Write the increment in specific values [+-x, +-y, 0] instead of placeholders. One relative position should only be listed once.
If you need to set distances or directions yourself, the Euclidean distances between objects should generally be greater than 1 meter, which is 1000 in coordinates, unless stated otherwise in the description.
2. For each relative position, if one object has a declared coordinate, calculate the coordinate of the other object. Then, list the positional information of each object, including coordinates and orientations.
You should first perform calculations to get the coordinates in the analysis part, then come up with the final positions.
3. For the remaining objects without coordinates, assign valid coordinates for them based on the restrictions of object position, direction, orientation, and relation.
The coordinate values should generally be from [-5000, 0] or [0, 5000] intervals unless stated otherwise in the description.
The Euclidean distances between objects should generally be greater than 1 meter, which is 1000 in coordinates, unless stated otherwise in the description.
All objects are on the ground with no vertical relative positions.
To avoid overlapping, coordinates assigned in this step should be different from existing objects and each other, with the exception of Guarding which surrounds the objects and whose position represents its center which is not occupied.
You should first perform calculations to get the coordinates and deduce whether the coordinates of all objects meet the above conditions in the analysis part, then come up with the final positions.
After this step, every object in the description should have a coordinate.
You should list the positional information, including coordinates and orientations, of all objects in Step 3.
You must not skip any step.
You should return in the following format:
```
#Step 1: Rewrite Relative Position#
Analysis: <analysis>
New Relative Positions:
[
    {
        "object 1": "<object name 1>",
        "relation": "[+-x, +-y, 0]",
        "object 2": "<object name 2>"
    },
    ...
]

#Step 2: Calculate Coordinates#
Analysis: <analysis and calculation>
Positions:
[
    {
        "name": "<object name>",
        "position": "[x, y, 0]",
        "orientation": "<orientation>"
    },
    ...
]

#Step 3: Assign Positions#
Analysis: <analysis and calculation>
Positions:
[
    {
        "name": "<object name>",
        "position": "[x, y, 0]",
        "orientation": "<orientation>"
    },
    ...
]
```
Do not say anything else.
\end{lstlisting}
\end{tcolorbox}

\vpara{Placement Verification}
\begin{tcolorbox}[left=0mm,right=0mm,top=0mm,bottom=0mm,boxsep=1mm,arc=0mm,boxrule=0pt, frame empty, breakable]
\small
\begin{lstlisting}
You are given a description of a workstation, which is used to build a scene in Process Simulate (PS).
The description includes several objects and their positions. The description is as follows:
```
{{ prompt }}
```
For directions and orientations, consider the front as the positive direction of the x-axis, the back as the negative direction of the x-axis, the left as the positive direction of the y-axis, and the right as the negative direction of the y-axis.
The orientation of an object is what direction the object faces, including an angle rotated counterclockwise on the ground in degrees or towards other objects. For example, rotating for 0 degrees is to face the front, 90 degrees is to face the left, 180 degrees is to face the back, and 270 degrees is to face the right. The default orientation is to rotate for 0 degrees which makes the object face the positive direction of the x-axis.
Based on the description, we allocate positions for the objects. The allocated positions of objects are as follows:
```
{{ placements }}
```
However, the positions may represent certain errors which need to be identified.
You should carry out the following actions to check whether there are conflicts in object positions:
1. Calculate each object's positional arrangement. In case objects' positions are described relatively, deduce the corresponding coordinates. All objects exist in a 2D plane with z-coordinate being 0.
2. Identify any existing errors among the positions of the objects:
   - Constraints: The description may provide additional positional constraints which the objects should not violate. The front is the positive direction of the x-axis and the left is the positive direction of the y-axis.
   - Conflicts: The position description may contain inconsistencies among the objects. The position of an object may be calculated in many ways, and if the results obtained from different calculations are not consistent, there are conflicts in the description.
   - Overlap: The Euclidean distances between objects should be greater than 1 meter, which is 1000 in coordinates, or the objects will overlap with each other. Note that one exception is the Guarding, which surrounds the objects inside it and whose position represents its center which is not occupied. Also note that objects can overlap with each other if the description requires them to do so, e.g. their coordinates are explicitly specified.
3. Report whether there are positional errors. If you cannot deduce whether there are any errors - for example, when precise positions cannot be computed, please report that there are no errors.
You should return in the following format:
```
Relations: <Calculate the coordinates and positional relations>
Analysis: <Based on the calculated positions, deduce whether there are errors>
Error: <Based on the analysis, decide whether there are errors in the description. Choose between "Yes" and "No". If you can not determine, choose "No">
```
Do not say anything else.
\end{lstlisting}
\end{tcolorbox}

\vpara{Placement Error Feedback}
\begin{tcolorbox}[left=0mm,right=0mm,top=0mm,bottom=0mm,boxsep=1mm,arc=0mm,boxrule=0pt, frame empty, breakable]
\small
\begin{lstlisting}
Your position allocation contains the following error:
```
{{ feedback }}
```
Please write the response again, fixing the errors.
Only generate the response. Do not say anything else.
\end{lstlisting}
\end{tcolorbox}

\vpara{Code Generation}
\begin{tcolorbox}[left=0mm,right=0mm,top=0mm,bottom=0mm,boxsep=1mm,arc=0mm,boxrule=0pt, frame empty, breakable]
\small
\begin{lstlisting}
You are given a description of a workstation, which is used to build a scene in Process Simulate (PS). You should write C# code with specific packages to build this scene.
The description is as follows:
```
{{ prompt }}
```
The objects that should be added into the scene are as follows: {{ objects }}
Their positions and orientations are as follows:
```
{{ placements }}
```
You should write the complete code in the following format:
```csharp
string rootDir = TxApplication.SystemRootDirectory;
string weldingLibPath = Path.Combine(rootDir, "Welding");
string[] weldingModels = Directory.GetDirectories(weldingLibPath, "*.cojt", SearchOption.TopDirectoryOnly);

/* create model list */

foreach (string model in weldingModels)
{
    DirectoryInfo directoryInfo = new DirectoryInfo(model);

    /* load models */
}

Random rand = new Random();
TxPhysicalRoot txPhysicalRoot = TxApplication.ActiveDocument.PhysicalRoot;

/* add objects into the scene */

TxApplication.RefreshDisplay();
```
where the parts surrounded by "/* */" are the parts you should fill.

For the "create model list" part, you should define Lists of DirectoryInfo, which are used to store models in the "add models" part. For example, if you want a list to store robots:
```csharp
List<DirectoryInfo> robotModels = new List<DirectoryInfo>();
```
You can name the model lists whatever you like in a similar form.

For the "load models" part, you should check whether the current `directoryInfo` belongs to the types of models that are needed in the scene one by one and add it to corresponding model list.
You should use the following methods to load the objects:
{{ guidance for loading object models }}

For the "add objects into the scene" part, you should add the objects into the scene and set its positions. For each object in the scene, pick the model from the model lists, put it into the scene, and set its coordinate and orientation.
To pick an object `obj1` from the list `objModels`, place it at [`x`, `y`, 0], and rotate it for `degree` degrees:
```csharp
DirectoryInfo objModel1 = objModels[rand.Next(0, objModels.Count)];
string obj1Name = Path.GetFileNameWithoutExtension(objModel1.Name) + "_" + DateTime.Now.ToString("yyyy-MM-dd-HH-mm-ss");
TxInsertComponentCreationData txInsertDataObj1 = new TxInsertComponentCreationData(obj1Name, objModel1.FullName);
ITxComponent txComponentObject1 = txPhysicalRoot.InsertComponent(txInsertDataObj1);

double transXValue1 = x;
double transYValue1 = y;
double rotValue1 = degree * Math.PI / 180.0;
TxTransformation txTransTransXYRotZ = new TxTransformation(new TxVector(transXValue1, transYValue1, 0.0), new TxVector(0.0, 0.0, rotValue1), TxTransformation.TxRotationType.RPY_ZYX);
ITxLocatableObject obj1 = (ITxLocatableObject)txComponentObject1;
obj1.AbsoluteLocation *= txTransTransXYRotZ;
```

Now, you should generate the complete code to build this scene. Only generate the complete code. Your code will be run directly in the production environment, so don't omit anything.
Please strictly follow the given code snippets to load and place objects. Do not generate a class or a function, instead directly generate the function body. Do not define or call custom classes or functions by yourself.
Please use .NET Framework 4.6.2 or below, or the code will not run.
Do not say anything else.
\end{lstlisting}
\end{tcolorbox}

\subsubsection{Ablation Study}
\vpara{Analyze Layout without \model}
\label{app:prompt-layout-analysis-single-call}
\begin{tcolorbox}[left=0mm,right=0mm,top=0mm,bottom=0mm,boxsep=1mm,arc=0mm,boxrule=0pt, frame empty, breakable]
\small
\begin{lstlisting}
You are given a description of a workstation, which is used to build a scene in Process Simulate (PS).
The original description is as follows:
```
{{ prompt }}
```
You should follow the steps below:
1. First, find all objects mentioned in the description and should appear in the workstation according to the description.
The objects that may appear in the description are from the permission list: [Kuka Robot KR125, Kuka Robot KR350, ABB Robot IRB6600, YASKAWA Robot ma01800, Welding Table, Turntable, Cabinet, ValveStand, Conveyor, Guarding]
If any object is not from the permission list or is from a general type, replace it with an object in the permission list that satisfies the restrictions in the description, including upper/lower cases.
If the description mentions multiple instances of the same object, list the object repeatedly for each instance, list the object name repeatedly for each instance and assign numbers in the object names to tell them apart. Note that a "station", a "workstation", or a "scene" is not an object.
You should return a list of objects in the following format:
```
Objects:
["ObjA", "ObjB 1", "ObjB 2", ...]
```
In the example, All objects in the list are chosen from the permission list and should appear in the scene according to the description. "ObjB 1" and "Obj 2" are two objects of the same kind. You should assign numbers to tell them apart.
2. Then, for each object in the object list generated in the previous step, set its placement in the scene according to the description. The placement includes coordinates and an orientations.
The coordinates are in the form of [x, y, 0], which should contain three values in millimeters in the form of "[x, y, 0]" with no unit after the brackets, with x and y representing the components of the coordinate on the x-axis and y-axis respectively.
To avoid overlapping, coordinates assigned in this step should be different from existing objects and each other, with the exception of Guarding which surrounds the objects and whose position represents its center which is not occupied.
The coordinate values should generally be from [-5000, 0] or [0, 5000] intervals unless stated otherwise in the description.
The Euclidean distances between objects should generally be greater than 1 meter, which is 1000 in coordinates, unless stated otherwise in the description.
All objects are on the ground with no vertical relative positions.
The orientation of an object is what direction the object faces, i.e. an angle rotated counterclockwise on the ground in degrees or towards other objects. For example, rotating for 0 degrees is to face the front, 90 degrees is to face the left, 180 degrees is to face the back, and 270 degrees is to face the right. The default orientation is to rotate for 0 degrees which makes the object face the positive direction of the x-axis.
For directions and orientations, consider the front as the positive direction of the x-axis, the back as the negative direction of the x-axis, the left as the positive direction of the y-axis, and the right as the negative direction of the y-axis.
You should first perform calculations to get the coordinates and orientations and deduce whether the placement of all objects satisfies the description in the analysis part, then come up with the final placement.
You should return a list of placements that appear in the description:
```
Placements:
[
    {
        "name": "<name>",
        "position": "<coordinate [x, y, 0]>",
        "orientation": "<orientation>"
    },
    ...
]
Use JSON format. <name> should be from the object list.
```

You should return your analysis, objects and placements in the format below:
```
#Step 1: Identify Objects#
Analysis: <analysis>
Objects:
[...]

#Step 2: Assign Placements#
Analysis: <analysis>
Placements:
[
    ...
]
```
Do not say anything else.
\end{lstlisting}
\end{tcolorbox}

\vpara{Code Generation without Object Placement API Guidance}
\label{app:prompt-code-gen-wo-placement-guidance}
\begin{tcolorbox}[left=0mm,right=0mm,top=0mm,bottom=0mm,boxsep=1mm,arc=0mm,boxrule=0pt, frame empty, breakable]
\small
\begin{lstlisting}
You are given a description of a workstation, which is used to build a scene in Process Simulate (PS). You should write C# code with Tecnomatix to build this scene.
The description is as follows:
```
{{ prompt }}
```
The objects that should be added into the scene are as follows: {{ objects }}
Their positions and orientations are as follows:
```
{{ placements }}
```
You should write the complete code in the following format:
```csharp
string rootDir = TxApplication.SystemRootDirectory;
string weldingLibPath = Path.Combine(rootDir, "Welding");
string[] weldingModels = Directory.GetDirectories(weldingLibPath, "*.cojt", SearchOption.TopDirectoryOnly);

/* create model list */

foreach (string model in weldingModels)
{
    DirectoryInfo directoryInfo = new DirectoryInfo(model);

    /* load models */
}

Random rand = new Random();
TxPhysicalRoot txPhysicalRoot = TxApplication.ActiveDocument.PhysicalRoot;

/* add objects into the scene */

TxApplication.RefreshDisplay();
```
where the parts surrounded by "/* */" are the parts you should fill.

For the "create model list" part, you should define Lists of DirectoryInfo, which are used to store models in the "add models" part. For example, if you want a list to store robots:
```csharp
List<DirectoryInfo> robotModels = new List<DirectoryInfo>();
```
You can name the model lists whatever you like in a similar form.

For the "load models" part, you should check whether the current `directoryInfo` belongs to the types of models that are needed in the scene one by one and add it to corresponding model list.
You should use the following methods to load the objects:
{{ guidance for loading object models }}

For the "add objects into the scene" part, you should add the objects into the scene and set its positions. For each object in the scene, pick the model from the model lists, put it into the scene, and set its coordinate and orientation.

Now, you should generate the complete code to build this scene. Only generate the complete code. Your code will be run directly in the production environment, so don't omit anything.
Please strictly follow the given code snippets to load and place objects. Do not generate a class or a function, instead directly generate the function body. Do not define or call custom classes or functions by yourself.
Please use .NET Framework 4.6.2 or below, or the code will not run.
Do not say anything else.
\end{lstlisting}
\end{tcolorbox}

\subsection{Use Of AI Assistants}
In this work, we use AI assistants to polish our writing.

\subsection{Benchmark}
\label{app:benchmark}
We list all the details of the industrial scene generation benchmark in Table~\ref{tab:benchmark-detail-1}, Table~\ref{tab:benchmark-detail-2}, and Table~\ref{tab:benchmark-detail-3}.

\begin{table*}[ht]
\centering
\caption{Benchmark details for descriptions with ID from 1 to 20.}
\label{tab:benchmark-detail-1}
\small
\begin{tabularx}{\textwidth}{c|X|l}
\toprule
\textbf{ID} & \textbf{Description} & \textbf{Category} \\ \midrule
1 & Create a welding station consisting of 3 robotic arms and a welding table. Arrange the 3 robotic arms in a straight line at equal intervals in front of the welding table. & Geometric Arrangement \\
2 & Can you design a virtual robotic welding cell setup? I'd like it to have a robot that picks up a part from two tables (at [1500, 0, 0], [-1500, 0, 0]) and moves part to a conveyor belt about 1.5m away. The conveyor should run parallel to the two tables. & Geometric Arrangement \\
3 & Create a layout featuring 6 robots positioned alongside two conveyor belts that are lined up end-to-end. Arrange 3 robots on each side of the conveyor belts in a straight line, with 1.5-meter intervals between each arm. & Geometric Arrangement \\
4 & Create a layout with 9 robots arranged in a 3x3 matrix configuration. Position the arms with a 3-meter interval between each one. & Geometric Arrangement \\
5 & How can we create a processing station where the robotic arm, work table, and conveyor belt are positioned at right angles to one another? Additionally, we need to enclose the area with protective guarding. & Geometric Arrangement \\
6 & Please set up a three-row layout. Place 3 work tables in the first row, position 3 robots in the second row, and install 2 conveyor belts in the third row. & Geometric Arrangement \\
7 & We need to set up an L-shaped scene. On one side, line up a series of tables in a row,spaced out by about 2 meter. On the other, put a row of cabinets, spaced out by about 1 meter. & Geometric Arrangement \\
8 & Provide me a U-shaped simulation layout. Position a row of worktables along the base. On one side, arrange four equally spaced cabinets. On the opposite side, install a series of conveyor belts. Place an ABB robot at the center of this layout. & Geometric Arrangement \\
9 & Picture a circular welding setup with a table in the middle and six robots around it. The robots can all reach the table easily, and there's a safety fence surrounding the whole thing. & Geometric Arrangement \\
10 & Create a welding workstation with a central work table. Position 4 KUKA robots on each side of the table, about 2 meters away. & Geometric Arrangement \\
11 & Create a welding workstation with a robot positioned at the center. Put 3 work tables and 1 conveyor belt around the robot, each placed in a different direction. & Geometric Arrangement \\
12 & Create a scenario with a Kuka Robot, a welding table, guarding, and a cabinet. Pick random types from the TX root 3D library. Space these components evenly in the scenario. & Geometric Arrangement \\
13 & Set up a U-shaped layout with 10 tables for the initial simulation scene. & Geometric Arrangement \\
14 & Position a Kuka robot in front of a table. & Geometric Arrangement \\
15 & Give me two conveyors, arranged in parallel. & Geometric Arrangement \\
16 & I need a simulation model for welding processes. Place an ABB robot at the center of the scene with a welding table 2 meter in front of it. Surround the entire scene with protective fencing. Additionally, position a KUKA robot arm at coordinates [1000, 2000, 0] for subsequent processing. & Positional Details \\
17 & I want a simulation scenario with the following layout: Position a welding table as the central element. Place a Kuka Robot KR125 2 meters in front of the table. To one side, at a distance of 4 meters from both the robot and table, arrange three cabinets in a row. Set the spacing between the first and second cabinet at 2.1 meters, and between the second and third cabinet at 1.3 meters. & Positional Details \\
18 & Create a simulation model with this layout: Position three ABB robots. Place the first robot at coordinates [2000, 2000, 0]. Align the other two robots along the y-axis of the first robot, with a 3-meter interval between each. Then, at a distance of 5 meters along the x-axis from the first robot, position a ValveStand next to a cabinet. & Positional Details \\
19 & I need a welding station with two KUKA robots for loading and unloading. The loading robot is at [2000, 2000, 0] and the unloading robot at [-2000, -2000, 0]. A conveyor belt transfers materials between loading and unloading positions. Additionally, two welding tables are situated near the loading and unloading robots. & Positional Details \\
20 & Please help me create a welding workstation including 1 Kuka robot and 1 ABB Robot, separately located 2m aside from a welding table in the center of the workstation. & Positional Details \\
\bottomrule
\end{tabularx}
\end{table*}

\newpage
\begin{table*}[ht]
\centering
\caption{Benchmark details for descriptions with ID from 21 to 33.}
\label{tab:benchmark-detail-2}
\small
\begin{tabularx}{\textwidth}{c|X|l}
\toprule
\textbf{ID} & \textbf{Description} & \textbf{Category} \\ \midrule
21 & Hey, could you set up a scene for me? I need an industrial welding setup with multiple models. I want a KUKA Robot KR125, a YASKAWA Robot, and 2 turntables. Can you put the KUKA at [2000, 0, 0], the YASKAWA at [0, 2000, 0], and one table at [-500, -500, 0], with another 2m behind it? & Positional Details \\
22 & Set up a welding station with a regular table, a welding table, two KUKA robots, a value stand, and safety guarding. Place one robot 1500mm from the regular table and the other 2000mm from the welding table. & Positional Details \\
23 & Put a pair of robots at the front of a new scene. Create a cabinet next to a turntable behind the scene. & Positional Details \\
24 & An ABB Robot is at [2000,2000]. A Table is at [25,20,0]. A Conveyor is at [-1234mm,-2001mm,0]. & Positional Details \\
25 & Create a scene with multiple robots and cabinets and 3 worktables. Position 3 worktables with a 2.5-meter interval between each. Ensure that every table is closely accompanied by two robots and one cabinet. & Object Quantity \\
26 & We'll need a table and five robots to set up an assembly line. Use three identical model robots, and two different models for the remaining two. & Object Quantity \\
27 & let's set up the scene with 5 ABB robots, position 3 welding tables next to 3 of the robots. Add 2 conveyor belts near the other robots. On one side of the scene, include 2 cabinets and one standard table. & Object Quantity \\
28 & Could you please generate a scene of a welding workstation with the following equipment: Kuka Robot KR350, ABB robot, guarding, and valve stands? Please apply a random number of each item in the scene. One important requirement is to position one of the Kuka Robots near one of the ABB robots, with a distance between them ranging from 4000 to 5000 millimeters. & Object Quantity \\
29 & I need to create a 3D model for a robotic welding cell. It's got a welding table in the middle and two Kuka KR125 robots around it, 3 abb robots 4 meters away. Can you help me set this up in PS. & Object Quantity \\
30 & The conveyor is positioned at coordinates [0,0,0]. 6 robots are uniformly distributed in front of the conveyor. & Object Quantity \\
31 & Design a complex welding station incorporating four robotic units: two ABB robots and two KUKA robots. Position the ABB robots at coordinates [2000,0,0] and [-2000,0,0], tasking them with product part loading and unloading operations. Place the KUKA robots at [0, 2000, 0] and [0, -2000,0], assigning them to perform collaborative spot welding tasks; add a guarding fence around the station, install a centrally positioned welding table and two conveyor systems to transport parts from the table. Additionally, place two equipment cabinets adjacent to the robots. & Composite Description \\
32 & These robotic arms are positioned in a triangular formation within the welding workstation, allowing them to work collaboratively on complex welding tasks. Each machine is equipped with advanced sensors and controllers for precise movement and coordination. The Kuka Robot KR350, located at [0, 0, 0], handles heavy-duty welding, while the ABB Robot IRB6600 at [2500, 2500, 0] assists with high-speed operations. The YASKAWA Robot MA01800 at [0, 2500, 0] focuses on intricate welding tasks. Together, these machines ensure high efficiency, accuracy, and flexibility for various welding operations. & Composite Description \\
33 & The proposed layout consists of a process simulation model featuring an U-shaped guarding structure, centrally located at the coordinates [1500, 1500, 0]. Inside the guarding, a square welding table and a cabinet are positioned, maintaining a 1.5-meter distance between them to ensure proper workspace and accessibility. The guarding serves to protect this working area, while allowing for efficient movement within the enclosed space.\newline Adjacent to the front side of the guarding, there is a bar-type conveyor system. Positioned alongside the conveyor is a valve stand and two robots, strategically positioned to enhance production flow. The overall design ensures a logical and organized layout, optimizing both safety and functionality within the workspace. & Composite Description \\
\bottomrule
\end{tabularx}
\end{table*}

\newpage
\begin{table*}[ht]
\centering
\caption{Benchmark details for descriptions with ID from 34 to 40.}
\label{tab:benchmark-detail-3}
\small
\begin{tabularx}{\textwidth}{c|X|l}
\toprule
\textbf{ID} & \textbf{Description} & \textbf{Category} \\ \midrule
34 & A welding workstation needs to be set up with three robotic manipulators and additional equipment. The main robots are:\newline Kuka Robot KR350 (Machine No. 1)\newline ABB Robot IRB6600 (Machine No. 2)\newline YASKAWA Robot MA01800 (Machine No. 3)\newline These robots are arranged in the following positions:\newline Machine No. 1: [0, 0, 0]\newline Machine No. 2: [2000,1500, 0]\newline Machine No. 3: [-2000, 2500, 0]\newline To complete the workstation, we need to add more equipment:\newline A conveyor belt is placed at [250, -1250, 0] to move materials.\newline A control cabinet is set up at [-200, -3250, 0] to manage the system.\newline A welding table is positioned at [0, 1500, 0] for the robots to work on.\newline A safety fence surrounds the entire area.\newline This setup creates a complete welding workstation with robots, material handling, and safety features. & Composite Description \\
35 & I need to create a simulation scenario with the following steps:\newline 1. Load a welding table\newline 2. Generate two robots beside it\newline 3. Protect the area with safety fencing\newline 4. Use ABB IRB6600 models for both robots\newline 5. Place two conveyor belts and three cabinets inside the fence\newline 6. Replace one of the robots with a KUKA model\newline 7. Remove one cabinet and position the remaining two next to the robots\newline 8. Ensure the welding table is at the center of the scene, then finalize the layout design & Composite Description \\
36 & Could you give an example of a real-world welding scenario in industry? & Fuzzy Description \\
37 & Build a welding workstation scene with 5 tables, and could you randomly add some typical items based on what you know about welding? & Fuzzy Description \\
38 & Please add robots, tables, and conveyors to create a welding scene. You may randomly determine the types, quantities, and spatial arrangements of these elements. & Fuzzy Description \\
39 & Hi, can you set up a welding scene for me? I need an initial environment for a welding use case, but without any robots. & Fuzzy Description \\
40 & Could you create an industrial welding workstation scene using the existing assets from our library? & Fuzzy Description \\
\bottomrule
\end{tabularx}
\end{table*}

\end{document}